\documentclass[a4paper]{article}
\usepackage[english]{babel}
\usepackage[utf8x]{inputenc}
\usepackage[T1]{fontenc}
\usepackage{times}
\newcommand{\mb}{\mathbf}
\usepackage{pgf,tikz,pgfplots}
\pgfplotsset{compat=1.14}
\usepackage{tabularx} 
\usepackage[round]{natbib}
\usepackage{amsmath}
\usepackage{algorithm}
\usepackage{algpseudocode}
\usepackage{amsfonts} 
\usepackage{soul}
\usepackage{bm}
\usepackage{amssymb}
\usepackage{multirow}
\usepackage{float}

\usepackage{afterpage}
\usepackage[colorinlistoftodos]{todonotes}
\usepackage{subcaption}
\usepackage{caption}

\usetikzlibrary{arrows.meta,
                positioning, calc, 
                shapes,chains, arrows, trees}

\definecolor{uququq}{rgb}{0.25098039215686274,0.25098039215686274,0.25098039215686274}
\tikzset{
    quote/.style={{|[width=2mm]}-{|[width=2mm]}}
}

\title{Assessing the Impact: Does an Improvement to a Revenue Management System Lead to an Improved Revenue?}

\date{}
\author{Greta Laage\footnote{Corresponding author. IVADO Labs and \'Ecole Polytechnique de Montr\'eal, Montr\'eal, Canada, {greta.laage@polymtl.ca}} \and Emma Frejinger \footnote{IVADO Labs, Canada Research Chair and DIRO, Universit\'e de Montr\'eal, Montr\'eal, Canada} \and Andrea Lodi\footnote{IVADO Labs, CERC and \'Ecole Polytechnique de Montr\'eal, Montr\'eal, Canada} \and Guillaume Rabusseau\footnote{IVADO Labs, Mila, Canada CIFAR AI Chair and DIRO, Universit\'e de Montr\'eal, Montr\'eal, Canada }}

\begin{document}

\maketitle

\vspace{0.5cm}

\begin{abstract}
Airlines and other industries have been making use of sophisticated Revenue Management Systems to maximize revenue for decades. While improving the different components of these systems has been the focus of numerous studies, estimating the impact of such improvements on the revenue has been overlooked in the literature despite its practical importance. Indeed, quantifying the benefit of a change in a system serves as support for investment decisions. 
This is a challenging problem as it corresponds to the difference between the generated value and the value that would have been generated keeping the system as before. The latter is not observable. Moreover, the expected impact can be small in relative value. 

In this paper, we cast the problem as counterfactual prediction of unobserved revenue.
The impact on revenue is then the difference between the observed and the estimated revenue. The originality of this work lies in the innovative application of econometric methods proposed for macroeconomic applications to a new problem setting. Broadly applicable, the approach benefits from only requiring revenue data observed for origin-destination pairs in the network of the airline at each day, before and after a change in the system is applied. 
We report results using real large-scale data from Air Canada.
We compare a deep neural network counterfactual predictions model with econometric models. They achieve respectively 1\% and 1.1\% of error on the counterfactual revenue predictions, and allow to accurately estimate small impacts (in the order of 2\%). 
\end{abstract}


\section{Introduction}

Airlines have been making use of sophisticated Revenue Management Systems (RMSs) to maximize revenue for decades. Through interacting prediction and optimization components, such systems handle demand bookings, cancellations and no-shows, as well as the optimization of seat allocations and overbooking levels. Improvements to existing systems are made by the airlines and solution providers in an iterative fashion, aligned with the advancement of the state-of-the-art where studies typically focus on one or a few components at a time \citep{talluri2006theory}. The development and maintenance of RMSs require large investments. In practice, incremental improvements are therefore often assessed in a proof of concept (PoC) prior to full deployment. The purpose is then to assess the performance over a given period of time and limited to certain markets, for example, a subset of the origin-destination pairs offered for the movement of passengers on the airline's network. We focus on a crucial question in this context: \emph{Does the improvement to the RMS lead to a significant improvement in revenue?} This question is difficult to answer because the value of interest is not directly observable. Indeed, it is the difference between the value generated during the PoC and \emph{the value that would have been generated} keeping business as usual. Moreover, the magnitude of the improvement can be small in a relative measure (for example, 1-3\%) while still representing important business value. Small relative values can be challenging to detect with statistical confidence.

Considering the wealth of studies aiming to improve RMSs, it is surprising that the literature focused on assessing quantitatively the impact of such improvements is scarce. We identify two categories of studies in the literature: First, those assessing the impact in a simplified setting leveraging simulation \citep[][]{weatherford2002,fiig2019}. These studies provide valuable information but are subject to the usual drawback of simulated environments. Namely, the results are valid assuming that the simulation behaves as the real system. This is typically not true for a number of reasons, for instance, assumptions on demand can be inaccurate and in reality there can be a human in the loop adjusting the system. Statistical field experiments do not have this drawback as they can be used to assess impacts in a real setting. Studies focusing on field experiments constitute our second category. There are, however, few applications in revenue management \citep{LopezEtAl21,KousEtAl12, PekgEtAl13} and even less focus on the airline industry \citep{cohen2018}. Each application presents its specific set of challenges. Our work can be seen as a field experiment whose aim is to assess if a PoC is a success or not with respect to a given success criteria. 

In practice, airlines often take a pragmatic approach and compare the value generated during a PoC to a simple baseline: either the revenue generated at the same time of the previous year, or the revenue generated by another market with similar behavior as the impacted market. This approach has the advantage of being simple. However, finding an adequate market is difficult, and the historical variation between the generated revenue and the baseline can exceed the magnitude of the impact that we aim to measure. In this case, the answer to the question of interest would be inconclusive.

We propose casting the problem as counterfactual prediction of the revenue without changing the RMS, and we compare it to the observed revenue generated during the PoC. Before providing background on counterfactual prediction models, we introduce some related vocabulary in the context of our application. Consider a sample of \emph{units} and observations of \emph{outcomes} for all units over a given time period. In our case, an example of a unit is an origin-destination (OD) pair and the observed outcome is the associated daily revenue. Units of interest are called \emph{treated units} and the other (untreated) units are referred to as \emph{control units}. In our case, the \emph{treatment} is a change to the RMS and it only impacts the treated units (in our example a subset of the ODs in the network). The goal is to estimate the \emph{untreated outcomes} of \emph{treated units} defined as a function of the outcome of the control units. In other words, the goal is to estimate what would have been the revenue for the treated OD pairs without the change to the RMS. We use the observed revenue of the untreated ODs for this purpose.

\paragraph{Brief background on counterfactual prediction models} 

\cite{doudchenko2016} and \cite{athey2018} review different approaches for imputing missing outcomes which include the three we consider for our application: (i) synthetic controls \citep{abadie2003, abadie2010} (ii) difference-in-differences \citep{ashenfelter1985, card1990impact, card2000minimum, athey2006identification}
and (iii) matrix completion methods \citep{mazumder2010spectral, candes2009exact, candes2010matrix}. \cite{doudchenko2016} propose a general framework for difference-in-differences and synthetic controls where the counterfactual outcome for the treated unit is defined as a linear combination of the outcomes of the control units.  
Methods (i) and (ii) differ by the constraints applied to the parameters of the linear combination. 
Those models assume that the estimated patterns across units are stable before and after the treatment
while models from the unconfoundedness literature \citep{imbens2015causal,rosenbaum1983central} estimate patterns from before treatment to after treatment
that are assumed stable across units. \cite{athey2018} qualify the former as vertical regression and the latter as horizontal regression. 
\cite{amjad2018} propose a robust version of synthetic controls based on de-noising the matrix of observed outcomes. 
\cite{poulos2017} proposes an alternative to linear regression methods, namely a non-linear recurrent neural network.  
\cite{athey2018} propose a general framework for counterfactual prediction models under matrix completion methods, 
where the incomplete matrix is the one of observed outcomes without treatment for all units at all time periods and the missing data patterns are not random. They draw on the literature on factor models and interactive fixed effects \citep{bai2003inferential, bai2002determining} where the untreated outcome is defined as the sum of a linear combination of covariates, that is, a low rank matrix and an unobserved noise component. 

The studies in the literature are mainly focused on macroeconomic applications. For example, estimating the economic impact on West Germany of the German reunification in 1990 \citep{AbadEtAl14}, the effect of a state tobacco control program on per capita cigarette sales \citep{abadie2010} and the effect of a conflict on per capita GDP \citep{abadie2003}. In comparison, our application exhibits some distinguishing features. First, the number of treated units can be large since airlines may want to estimate the impact on a representative subset of the network. Often there are hundreds, if not thousands of ODs in the network. Second, the number of control units is potentially large but the network structure leads to potential spillover effects that need to be taken into account. Third, even if the number of treated units can be large, the expected treatment effect is typically small. In addition, airline networks are affected by other factors, such as weather and seasonality. Their impact on the outcome needs to be disentangled from that of the treatment.

\paragraph{Contributions} This paper offers three main contributions. First, we formally introduce the problem and provide a comprehensive overview of existing counterfactual prediction models that can be used to address it. Second, based on real data from Air Canada, we provide an extensive computational study showing that the counterfactual predictions accuracy is high when predicting revenue. We focus on a setting with multiple treated units and a large set of controls. We present a non-linear deep learning model to estimate the missing outcomes that takes as input the outcome of control units as well as time-specific features. The deep learning model achieves less than 1\% error for the aggregated counterfactual predictions over the treatment period. Third, we present a simulation study of treatment effects showing that we can accurately estimate the effect even when it is relatively small.

\paragraph{Paper Organization.} The remainder of the paper is structured as follows. Next we present a thorough description of the problem. We describe in Section~\ref{section:existing_methods} the different counterfactual prediction models. 
In Section~\ref{section:experimental_setting}, we describe our experimental setting and the results of an extensive computational study.
Finally, we provide some concluding remarks in Section~\ref{section:conclusion}.

\section{Problem Description}
\label{section:problem_description}

In this section, we provide a formal description of the problem and follow closely the notation from \cite{doudchenko2016} and \cite{athey2018}.

We are in a panel data setting with $N$ units covering time periods indexed by $t=1, \ldots, T$. A subset of units is exposed to a binary treatment during a subset of periods. We observe the realized outcome for each unit at each period. In our application, a unit is an OD pair and the realized outcome is the \textit{booking issue date revenue} at time $t$, that is, the total revenue yielded at time $t$ from bookings made at $t$. The methodology described in this paper is able to handle various types of treatments, assuming it is applied to a subset of units. 
The set of \textit{treated units} receive the treatment and the set of \textit{control units} are not subject to any treatment. The treatment effect is the difference between the observed outcome under treatment and the outcome without treatment. The latter is unobserved and we focus on estimating the missing outcomes of the treated units during the treatment period. 

We denote $T_0$ the time when the treatment starts and split the complete observation period into a pre-treatment period $t = 1, \ldots, T_0 $ and a treatment period $t = T_0+1, \ldots, T$. We denote $T_1 = T-T_0$ the length of the treatment period. Furthermore, we partition the set of units into treated $i = 1, \ldots, N^{\text{t}}$ and control units $i = N^{\text{t}} + 1, \ldots, N$, where the number of control units is $N^{\text{c}} = N - N^{\text{t}}$.

In the pre-treatment period, both control units and treated units are untreated. In the treatment period, only the control units are untreated and, importantly, we assume that they are unaffected by the treatment. The set of treated pairs $(i, t)$ is
\begin{equation}
    \label{eq:treated_it}
    \mathcal{M} = \{ (i, t) \text{ } i = 1,\ldots,N^{\text{t}}, t = T_0+1,\ldots,T \},
\end{equation}
and the set of untreated pairs $(i, t)$ is
\begin{equation}
    \label{eq:untreated_it}
    \mathcal{O} = \{ (i, t) \text{ } i = 1,\ldots,N^{\text{t}}, t = 1,\ldots,T_0 \} \cup \{ (i, t) \text{ } i = N^{\text{t}}+1,\ldots,N, t = 1,\ldots,T \}.
\end{equation}
Moreover, the treatment status is denoted by $W_{it}$ and is defined as
\begin{equation}
  W_{it}=\left\{
  \begin{array}{@{}ll@{}}
    1 & \text{ if } (i, t) \in \mathcal{M} \\
    0 & \text{ if } (i, t) \in \mathcal{O}.
  \end{array}\right.
\end{equation} 

For each unit $i$ in period $t$, we observe the treatment status $W_{it}$ and the realized outcome $Y_{it}^{\text{obs}} = Y_{it}(W_{i t})$. Our objective is to estimate $\hat{Y}_{it}(0)~\forall (i,t) \in \mathcal{M}$. Counterfactual prediction models define the latter as a mapping of the outcome of the control units.

The observation matrix, denoted by $\mathbf{Y}^{\text{obs}}$ is a $N \times T$ matrix whose components are the observed outcomes for all units at all periods. The first $N^{\text{t}}$ rows correspond to the outcomes for the treated units and the first $T_0$ columns to the pre-treatment period. The matrix $\mathbf{Y}^{\text{obs}}$ hence has a block structure,
\begin{equation}
\nonumber
    \mb{Y}^{\text{obs}} = \begin{pmatrix}
\mb{Y}_{\text{pre}}^{\text{obs,t}} & \mb{Y}_{\text{post}}^{\text{obs,t}} \\
\mb{Y}_{\text{pre}}^{\text{obs,c}} & \mb{Y}_{\text{post}}^{\text{obs,c}}
\end{pmatrix},
\end{equation}
where $\mb{Y}_{\text{pre}}^{\text{obs,c}}$ (respectively $\mb{Y}_{\text{pre}}^{\text{obs,t}}$) represents the $N^{\text{c}} \times T_0$ (resp. $N^{\text{t}} \times T_0$) matrix of observed outcomes for the control units (resp. treated units) before treatment. Similarly, $\mb{Y}_{\text{post}}^{\text{obs,c}}$ (respectively $\mb{Y}_{\text{post}}^{\text{obs,t}}$) represents the $N^{\text{c}} \times T_1$ (resp. $N^{\text{t}} \times T_1$) matrix of observed outcomes for the control units (resp. treated units) during the treatment.

Synthetic control methods have been developed to estimate the average causal effect of a treatment \citep{abadie2003}. Our focus is slightly different as we aim at estimating the total treatment effect during the treatment period $T_0+1,\ldots,T$,
\begin{equation}
    \tau = \sum_{i = 1}^{N^{\text{t}}} \sum_{t = T_0+1}^T Y_{it}(1) - Y_{it}(0).
    \label{eq:total_treatment}
\end{equation}

We denote by $\hat{\tau}$ the estimated treatment effect,
\begin{equation}
    \hat{\tau} = \sum_{i = 1}^{N^{\text{t}}} \sum_{t = T_0+1}^T Y_{it}^{\text{obs}} - \hat{Y}_{it}(0). 
    \label{eq:impact}
\end{equation}

\section{Counterfactual Prediction Models}
\label{section:existing_methods}

In this section, we describe counterfactual prediction models from the literature that can be used to estimate the missing outcomes $Y_{it}(0) ~\forall (i, t) \in \mathcal{M}$. Namely, grouped under synthetic control methods (Section~\ref{sec:litt_syntheticControl}), we describe the constrained regressions in \cite{doudchenko2016} which include difference-in-differences and synthetic controls from \cite{abadie2010}. In Section~\ref{sec:litt_robustSynthetic}, we delineate the robust synthetic control estimator from \cite{amjad2018} followed by the matrix completion with nuclear norm minimization from \cite{athey2018} in Section~\ref{sec:litt_nuclearnorm}. 
Note that we present all of the above with one single treated unit, i.e., $N^{\text{t}} = 1$. This is consistent with our application as we either consider the units independently, or we sum the outcome of all treated units to form a single one.
Finally, in Section~\ref{sec:FFNN}, we propose a feed-forward neural network architecture that either considers a single treated unit or several to relax the independence assumption.

\subsection{Synthetic Control Methods} \label{sec:litt_syntheticControl}

\cite{doudchenko2016} propose the following linear structure for  estimating the unobserved $Y_{it}(0)$, $(i,t) \in \mathcal{M}$, arguing that several methods from the literature share this structure. More precisely, it is a linear combination of the control units,
\begin{equation}
    Y_{it}(0) = \mu + \sum_{j = N^{\text{t}}+1}^N \omega_j Y_{j t}^{\text{obs}} + e_{it} \quad \forall (i, t) \in \mathcal{M},
    \label{eq:LR_counterfactuals}
\end{equation}
where $\mu$ is the intercept, $\boldsymbol{\omega} = (\omega_1, \ldots, \omega_{N^{\text{c}}})^{\top}$ a vector of $N^{\text{c}}$ parameters and $e_{it}$ an error term. 

Synthetic control methods differ in the way the parameters of the linear combination are chosen depending on specific constraints and the observed outcomes $\mb{Y}_{\text{pre}}^{\text{obs,t}}$, $\mb{Y}_{\text{pre}}^{\text{obs,c}}$ and $\mb{Y}_{\text{post}}^{\text{obs,c}}$. We write it as an optimization problem with an objective function minimizing the sum of least squares
\begin{equation}
    \min_{\mu, \boldsymbol{\omega}} \left\Vert  \mb{Y}_{\text{pre}}^{\text{obs,t}} - \mu \mathbf{1}_{T_0}^{\top} - \boldsymbol{\omega}^{\top} \mb{Y}_{\text{pre}}^{\text{obs,c}} \right\Vert ^2,  \label{eq:objective}
\end{equation}
potentially subject to one or several of the following constraints
\begin{align}
 & \quad \mu = 0 \label{eq:no_intercept}\\
    & \sum_{j = N^{\text{t}}+1}^N \omega_j = 1  \label{eq:adding_up}\\
    & \omega_j \geq 0 , \quad j = N^{\text{t}}+1,\ldots,N \label{eq:no_neg} \\
    & \omega_j = \Bar{\omega}, \quad j = N^{\text{t}}+1,\ldots,N. \label{eq:cst_weights}
\end{align}

In the objective \eqref{eq:objective}, $\mathbf{1}_{T_0}$ denotes a $T_0$ vector of ones. Constraint (\ref{eq:no_intercept}) enforces no intercept and  (\ref{eq:adding_up}) constrains the sum of the weights to equal one.  Constraints (\ref{eq:no_neg}) impose non-negative weights. 
Finally, constraints (\ref{eq:cst_weights}) force all the weights to be equal to a constant. 
If $T_0 \gg N$, \cite{doudchenko2016} argue that 
the parameters $\mu$ and $\boldsymbol{\omega}$ can be estimated by least squares, without any of the constraints (\ref{eq:no_intercept})-(\ref{eq:cst_weights}) and we may find a unique solution $(\mu, \boldsymbol{\omega})$. As we further detail in Section~\ref{section:experimental_setting}, this is the case in our application. We hence ignore all the constraints and estimate the parameters by least squares.

\subsubsection{Difference-in-Differences}

The Difference-In-Differences (DID) methods \citep{ashenfelter1985, card1990impact, card2000minimum, meyer1995workers, angrist1999empirical, bertrand2004much, angrist2008mostly, athey2006identification} consist in solving 
\begin{align*}
    (\textit{DID}) \quad \min_{\mu, \boldsymbol{\omega}} & \left\Vert  \mb{Y}_{\text{pre}}^{\text{obs,t}} - \mu \mathbf{1}_{T_0}^{\top} - \boldsymbol{\omega}^{\top} \mb{Y}_{\text{pre}}^{\text{obs,c}} \right\Vert ^2  \tag{\ref{eq:objective}}\\
    \text{s.t. } & \text{(\ref{eq:adding_up}), (\ref{eq:no_neg}), (\ref{eq:cst_weights})}. \nonumber
\end{align*}{}
With one treated unit and $N^{\text{c}} = N-1$ control units, solving $(\textit{DID})$ leads to the following parameters and counterfactual predictions:
\begin{align}
    & \hat{\omega}_j^{\text{DID}} = \frac{1}{N-1} , \quad j = 2, \ldots, N\\
    & \hat{\mu}^{\text{DID}} = \frac{1}{T_0} \sum_{t = 1}^{T_0} Y_{1 t} - \frac{1}{(N-1)  T_0} \sum_{t = 1}^{T_0} \sum_{j = 2}^{N} Y_{jt} \\
    & \hat{Y}_{1t}^{\text{DID}} (0) = \hat{\mu}^{\text{DID}} + \sum_{j = 2}^N \hat{\omega}_j^{\text{DID}} Y_{jt}.
\end{align}

\subsubsection{Abadie-Diamond-Hainmueller Synthetic Control Method}
\label{subsection:adh_sc}
Introduced in \cite{abadie2003} and \cite{abadie2010}, the synthetic control approach consists in solving 
\begin{align*}
    \textit{(SC)} \quad \min_{\mu, \boldsymbol{\omega}} & \left\Vert  \mb{Y}_{\text{pre}}^{\text{obs,t}} - \mu \mathbf{1}_{T_0}^{\top} - \boldsymbol{\omega}^{\top} \mb{Y}_{\text{pre}}^{\text{obs,c}} \right\Vert ^2  \tag{\ref{eq:objective}}\\
    \text{s.t. } & \text{(\ref{eq:no_intercept}), (\ref{eq:adding_up}), (\ref{eq:no_neg})}. \nonumber
\end{align*}{}
Constraints \eqref{eq:no_intercept}, \eqref{eq:adding_up} and \eqref{eq:no_neg} enforce that the treated unit is defined as a convex combination of the control units with no intercept. 

The (SC) model is challenged in the presence of non-negligible levels of noise and missing data in the observation matrix $\mb{Y}^{\text{obs}}$. Moreover, it is originally defined for a small number of control units and relies on having deep domain knowledge to identify the controls.

\subsubsection{Constrained Regressions}
The estimator proposed by \cite{doudchenko2016}  consists in solving
\begin{equation}
    \textit{(CR-EN)} \quad \min_{\mu, \boldsymbol{\omega}} \left\Vert  \mb{Y}_{\text{pre}}^{\text{obs,t}} - \mu \mathbf{1}_{T_0}^{\top} - \boldsymbol{\omega}^{\top} \mb{Y}_{\text{pre}}^{\text{obs,c}} \right\Vert_2^2 + \lambda^{\text{CR}} \left( \frac{1 - \alpha^{\text{CR}}}{2} ||\boldsymbol{\omega}||_2^2  + \alpha^{\text{CR}} ||\boldsymbol{\omega}||_1 \right),
    \label{eq:doudchenko}
\end{equation}{}
while possibly imposing a subset of the constraints (\ref{eq:no_intercept})-(\ref{eq:cst_weights}). 

The second term of the objective function (\ref{eq:doudchenko}) serves as regularization. This is an \textit{elastic-net} regularization that combines the Ridge term which forces small values of weights and Lasso term which reduces the number of weights different from zero. It requires two parameters $\alpha^{\text{CR}}$ and $\lambda^{\text{CR}}$. To estimate their values, the authors propose a cross-validation procedure, where each control unit is alternatively considered as a treated unit and the remaining control units keep their role of control. They are used to estimate the counterfactual outcome of the treated unit. The parameters chosen minimize the mean-squared-error (MSE) between the estimations and the ground truth (real data) over the $N^{\text{c}}$ validations sets.

The chosen subset of constraints depends on the application and the ratio of the number of time periods over the number of control units. In our experimental setting, we have a large number of pre-treatment periods, i.e., $T_0 \gg N^{\text{c}}$ and we focus on solving \textit{(CR-EN)} without constraints.

\subsection{Robust Synthetic Control} \label{sec:litt_robustSynthetic}

To overcome the challenges of $(SC)$ described in Section~\ref{subsection:adh_sc}, \cite{amjad2018} propose the Robust Synthetic Control algorithm. It consists in two steps: The first one de-noises the data 
and the second step learns a linear relationship between the treated units and the control units under the de-noising setting. The intuition behind the first step is that the observation matrix contains both the valuable information and the noise. The noise can be discarded when the observation matrix is approximated by a low rank matrix, estimated with singular value thresholding  \citep{chatterjee2015matrix}. Only the singular values associated with valuable information are kept. 
The authors posit that for all units without treatment,
\begin{equation}
    Y_{it}(0) = M_{it} + \epsilon_{it}, \quad i = 1, \ldots, N ,\quad t = 1,\ldots,T, 
    \label{eq:rsc_model}
\end{equation}
where $M_{it}$ is the mean and $\epsilon_{it}$ is a zero-mean noise independent across all $(i, t)$ (recall that for $ (i, t) \in \mathcal{O}$, $Y_{i t}(0) = Y_{i t}^{\text{obs}}$). 
A key assumption is that a set of weights $\{\beta_{N^{\text{t}} +1}, \ldots, \beta_N \}$ exist such that 
\begin{equation}
    M_{it} = \sum_{j = N^{\text{t}}+1}^N \beta_j M_{j t}, \quad i = 1, \ldots, N^{\text{t}}, \quad t = 1, \ldots,T.
\end{equation}

Before treatment, for $t\leq T_0$, we observe $Y_{i t}(0)$ for all treated and control units. In fact, we observe $M_{it}(0)$ with noise. 
The latent matrix of size $N \times T$ is denoted $\mathbf{M}$. We follow the notation in Section~\ref{section:problem_description}: $\mb{M}^{\text{c}}$ is the latent matrix of control units and $\mb{M}_{\text{pre}}^{\text{c}}$ the latent matrix of the control units in the pre-treatment period. We denote $\hat{\mathbf{M}}^{\text{c}}$ the estimate of $\mb{M}^{\text{c}}$ and $\hat{\mb{M}}_{\text{pre}}^{\text{c}}$ the estimate of $\mb{M}_{\text{pre}}^{\text{c}}$. With one treated unit, $i = 1$ designates the treated unit 
and the objective is to estimate $\hat{\mb{M}}^{\text{t}}$, the latent vector of size $T$ of treated units. The two-steps algorithm is described in Algorithm~\ref{algo:RSC}. It takes two hyperparameters: the singular value threshold $\gamma$ and the regularization coefficient $\eta$.

\begin{algorithm}[H]
\caption{Robust Synthetic Control \citep{amjad2018}}
\label{algo:RSC}
\begin{algorithmic}[1]
\State \textbf{Input}: $\gamma$, $\eta$
\State \textbf{Step 1}: De-noising the data with singular value threshold
\State \hspace{1cm}  Singular value decomposition of $\mathbf{Y}^{\text{obs,c}}$:  $\mathbf{Y}^{\text{obs,c}} = \sum_{i = 2}^{N} s_i u_i v_i^{\top}$
\State \hspace{1cm}  Select the set of singular values above $\gamma$ : $S = \{ i: s_i \geq \gamma \}$
\State \hspace{1cm}  Estimator $\hat{\mathbf{M}}^{\text{c}} = \frac{1}{\hat{p}} \sum_{i \in S} s_i u_i v_i^{\top}$, where $\hat{p}$ is the fraction of observed data
\State \textbf{Step 2}: Learning the linear relationship between controls and treated units
\State \hspace{1cm} $\hat{\boldsymbol{\beta}}(\eta) = \arg\min_{\mathbf{b} \in \mathbb{R}^{N-1}} \left\Vert \mb{Y}_{\text{pre}}^{\text{obs,t}} -\hat{\mb{M}}_{\text{pre}}^{\text{c}\top} \boldsymbol{b} \right\Vert^2$
+ $\eta || \boldsymbol{b} ||_2^2$.
\State \hspace{1cm} Counterfactual means for the treatment unit: $\hat{\mb{M}}^{\text{t}} = \hat{\mathbf{M}}^{\text{c}\top} \hat{\boldsymbol{\beta}}(\eta)$
\State \textbf{Return $\hat{\boldsymbol{\beta}}$} :
\begin{equation}
    \hat{\boldsymbol{\beta}}(\eta) = \left( \hat{\mb{M}}_{\text{pre}}^{\text{c}}  (\hat{\mb{M}}_{\text{pre}}^{\text{c}\top} + \eta \mathbf{I} )\right)^{-1} \hat{\mb{M}}_{\text{pre}}^{\text{c}} \mb{Y}_{\text{ pre}}^{\text{t}}
\end{equation}
\end{algorithmic}
\end{algorithm}

\cite{amjad2018} prove that the first step of the algorithm (which de-noises the data) allows to obtain a consistent estimator of the latent matrix. Hence, the estimate $\hat{\mathbf{M}}^{\text{c}}$ obtained with Algorithm~\ref{algo:RSC} is a good estimate of $\mathbf{M}^{\text{c}}$ when the latter is low rank.

The threshold parameter $\gamma$ acts as a way to trade-off the bias and the variance of the estimator. Its value can be estimated with cross-validation. 
The regularization parameter $\eta \geq 0$ controls the model complexity. To select its value, the authors recommend to take the forward chaining strategy, which maintains 
the temporal aspect of the pre-treatment data. It proceeds as follows. For each $\eta$, for each $t$ in the pre-treatment period, split the data into 2 sets: $1, \ldots, t-1$ and $t$, where the last point serves as validation and select as value for $\eta$ the one that minimizes the MSE averaged over all validation sets.

\subsection{Matrix Completion with Nuclear Norm Minimization} \label{sec:litt_nuclearnorm}

\cite{athey2018} propose an approach inspired by matrix completion methods. They posit a model similar to \eqref{eq:rsc_model}, 
\begin{equation}
Y_{it}(0) = L_{i t} + \varepsilon_{i t}, \quad i=1,\ldots,N, \quad t=1,\ldots,T, 
\end{equation}
where $\varepsilon_{i t}$ is a measure of error. This means that during the pre-treatment period, we observe $L_{i t}$ with some noise. The objective is to estimate the $N \times T$ matrix $\mb{L}$. \cite{athey2018} assume that the matrix $\mathbf{L}$ is low rank and hence can be estimated with a matrix completion technique. The estimated counterfactual outcomes of treated units without treatment $\hat{Y}_{i t}(0), (i, t) \in \mathcal{M}$ is given by the estimate ${\hat{L}_{i t}}, (i, t) \in \mathcal{M}$. 

We use the following notation from \cite{athey2018} to introduce their estimator. For any matrix $\mb{A}$ of size $N \times T$ with missing entries  $\mathcal{M}$ and observed entries $\mathcal{O}$,  $P_{\mathcal{O}}(\mathbf{A})$ designates the matrix with values of $\mathbf{A}$, where the missing values are replaced by 0 and $P_{\mathcal{O}}^{\perp}(\mathbf{A})$ the one where the observed values are replaced by 0. 

They propose the following estimator of $\mathbf{L}$ from \cite{mazumder2010spectral}, for a fixed value of $ \lambda^{\text{mc}}$, the regularization parameter:
\begin{equation}
    \hat{\mathbf{L}} = \arg\min_{\mathbf{L}} \left \{ \frac{1}{|\mathcal{O}|} || P_{\mathcal{O}}(\mathbf{Y}^{\text{obs}}- \mathbf{L})||_F^2 +  \lambda^{\text{mc}} ||\mathbf{L}||_* \right\} , 
    \label{eq:mcnnm_obj}
\end{equation}
where $|| \mathbf{L} ||_F$ is the Fröbenius norm defined by 
\begin{equation}
    || \mathbf{L} ||_F = \left( \sum_{i} \sigma_i(\mathbf{L})^2 \right)^2 = \left( \sum_{i = 1}^N \sum_{t = 1}^T L_{i t}^2 \right)^2
\end{equation}
with $\sigma_i$ the singular values and $||\mathbf{L}||_*$ is the nuclear norm such that $ || \mathbf{L} ||_* = \sum_i \sigma_i(\mathbf{L}) $. The first term of the objective function~\eqref{eq:mcnnm_obj} is the distance between the latent matrix and the observed matrix. The second term is a regularization term encouraging $\mathbf{L}$ to be low rank. 

\cite{athey2018} show that their proposed method and synthetic control approaches are matrix completion methods based on matrix factorization. They rely on the same objective function which contains the Fröbenius norm of the difference between the unobserved and the observed matrices. Unlike synthetic controls that impose different sets of restrictions on the factors, they only use regularization. 

\cite{athey2018} use the convex optimization program SOFT-IMPUTE from \cite{mazumder2010spectral} described in Algorithm \ref{algo:MCNNM} to estimate the matrix $\mathbf{L}$. 
With the singular value decomposition $\mb{L} = \mathbf{S \mathbf{\Sigma} R}^{\top}$, the matrix shrinkage operator is defined by $\text{shrink}_{\lambda^{\text{mc}}} (\mb{L}) = \mathbf{S \Tilde{\mathbf{\Sigma}} R}^{\top}$, where $\Tilde{\mathbf{\Sigma}}$ is equal to $\mathbf{\Sigma}$ with the $i$-th singular value  replaced by $\max(\sigma_i(\mb{L}) - \lambda^{\text{mc}}, 0)$.

\begin{algorithm}[htbp]
\caption{SOFT-IMPUTE \citep{mazumder2010spectral} for Matrix Completion with Nuclear Norm Maximization \citep{athey2018}}
\label{algo:MCNNM}
\begin{algorithmic}[1]
\State \textbf{Initialization}: $\mb{L}_1( \lambda^{\text{mc}}, \mathcal{O}) = \mathbf{P}_{\mathcal{O}}(\mathbf{Y}^{\text{obs}})$
\For{$k = 1$ until $\{ \mb{L}_{k}( \lambda^{\text{mc}}, \mathcal{O}) \}_{k\geq1}$ converges}
    \State $\mb{L}_{k+1}( \lambda^{\text{mc}}, \mathcal{O}) = \text{shrink}_{\frac{ \lambda^{\text{mc}} |\mathcal{O}|}{2}} (\mathbf{P}_{\mathcal{O}}(\mathbf{Y}^{\text{obs}}) +  \mathbf{P}_{\mathcal{O}}^{\perp}\left( \mb{L}_k(\lambda) \right)) $
\EndFor
\State $\hat{\mb{L}}( \lambda^{\text{mc}}, \mathcal{O}) = \lim_{k\to\infty} \mb{L}_k( \lambda^{\text{mc}}, \mathcal{O})$
\end{algorithmic}
\end{algorithm}

The value of $ \lambda^{\text{mc}}$ can be selected via cross-validation as follows: For $K$ subsets of data among the observed data with the same proportion of observed data as in the original observation matrix, for each potential value of $ \lambda_j^{\text{mc}}$, compute the associated estimator $\hat{\mb{L}}( \lambda_j^{\text{mc}}, \mathcal{O}_k)$ and the MSE on the data without $\mathcal{O}_k$. Select the value of $\lambda$ that minimizes the MSE.  To fasten the convergence of the algorithm, the authors recommend to use $\hat{\mb{L}}(\lambda_j^{\text{mc}}, \mathcal{O}_k)$ as initialization for $\hat{\mb{L}}(\lambda_{j+1}^{\text{mc}}, \mathcal{O}_k)$ for each $j$ and $k$.

\subsection{Feed-forward Neural Network} \label{sec:FFNN}

In this section, we propose a deep learning model to estimate the missing outcomes and detail the training of the model. We consider two possible configurations: (i) when there is one treated unit and (ii) when there are multiple dependent treated units. In (i), the output layer of the model has one neuron. In (ii), the output layer contains $N^\text{t}$ neurons. The model learns the dependencies between treated units and predicts simultaneously the revenue for all of them. 

We define the counterfactual outcomes of the treated units as a non-linear function $g$ of the outcomes of the control units with parameters $\theta^{\text{ffnn}}$ and matrix of covariates $\mathbf{X}$ 

\begin{equation}
    \mathbf{Y}^{\text{t}}(0) = g\left( \mathbf{Y}^{\text{obs,c}}, \mathbf{X}, \theta^{\text{ffnn}} \right).
    \label{eq:nonLR_counterfactuals}
\end{equation}

In the following subsections, we use terminology from the deep learning literature \citep[][]{Goodfellow2016} but keep the notations described in Section~\ref{section:problem_description}. We define $g$ to be a feed-forward neural network (FFNN) architecture. We describe next the architecture in detail along with the training procedure. 

\subsubsection{Architecture}

\cite{barron1994approximation} shows that multilayer perceptrons (MLPs), also called FFNNs, are considerably more efficient than linear basis functions to approximate smooth functions. When the number of inputs $I$ grows, the required complexity for an MLP only grows as $\mathcal{O}(I)$, while the complexity for a linear basis function approximator grows exponentially for a given degree of accuracy. 
When $N^{\text{t}} > 1$, the architecture is multivariate, i.e., the output layer has multiple neurons. It allows parameter sharing between outputs and thus considers the treated units as dependent.

Since historical observations collected prior to the beginning of the treatment period are untreated, the counterfactual prediction problem can be cast as a supervised learning problem on the data prior to treatment. The features are the observed outcomes of the control units and the targets are the outcomes of the treated units. 
The pre-treatment period is used to train and validate the neural network and the treatment period forms the test set. This is a somewhat unusual configuration for supervised learning. Researchers usually know the truth on the test set also and use it to evaluate the ability to generalize. To overcome this difficulty, we describe in Section~\ref{subsection:seq_valid} a sequential validation procedure that aims at mimicking the standard decomposition of the dataset into training, validation and test sets. 

\begin{figure}
\centering
\begin{tikzpicture}[ 
node distance = 0mm and 5mm,
    block/.style = {draw, rounded corners, minimum height=6em, text width=5em, align=center},
    block2/.style = {draw, rounded corners, minimum height=6em, text width=4em, align=center},
    halfblock/.style = {block, minimum height=2em},
    halfblock2/.style = {block, minimum height=2em},
     line/.style = {draw, -Latex}
                       ]
\node (n1) [block]  {HIDDEN FC LAYERS};
\node (n2) [block2, right=of n1] {OUTPUT \\ LAYER};  library
\node (n1a) [halfblock, below left=of n1.north west]  {FC \\ LAYER};
\node (n1b) [halfblock2, above left=of n1.south west]  {FC \\ LAYER};
\node (n1a1) [halfblock, left=of n1a]  {INPUT LAYER 1};
\node (n1b1) [halfblock, left=of n1b]  {INPUT LAYER 2};
\path [line]    (n1)  edge (n2)
                (n1a) edge (n1.west |- n1a)
                (n1b)  --  (n1.west |- n1b)
                (n1a1) edge (n1a)
                (n1b1) edge (n1b);
\end{tikzpicture}
\caption{FFNN Architecture with Fully Connected (FC) layers.}
\label{fig:nn_draw}
\end{figure}
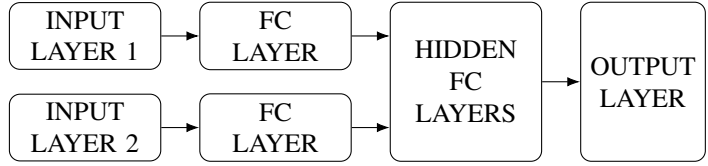

We present in Figure~\ref{fig:nn_draw} the model architecture. We use  two input layers to differentiate features. Input Layer 1 takes external features, and Input Layer 2 takes the lagged outcomes of control units. Let us consider the prediction at day $t$ as illustration. When $t$ is a day, it is associated for instance to a day of the week $\textit{dow}_t$, a week of the year $\textit{woy}_t$ and a month $m_t$. The inputs at Input Layer 1 could then be $\textit{dow}_t, \textit{woy}_t, m_t$.  Lagged features of control units are $Y_{it'}, i = N^{\text{t}}+1, \ldots, N$ and $t' = t, t-1, \ldots, t-l,$ where $l$ is the number of lags considered. They are fed into Input Layer 2. The output layer outputs $N^{\text{t}}$ values, one for each treated unit.

\subsubsection{Sequential Validation Procedure and Selection of Hyper-parameters}
\label{subsection:seq_valid}

In standard supervised learning problems, the data is split into training, validation and test datasets, where the validation dataset is used for hyper-parameters search. Table~\ref{table:hyperparams} lists the hyper-parameters of our architecture and learning algorithm. For each potential set of hyper-parameters $\Theta$, the model is trained on the training data and we estimate the parameters $\theta^{\text{ffnn}}$. We compute the MSE between the predictions and the truth on the validation dataset. We select the set $\Theta$ which minimizes the MSE. 

\begin{table}[H]
\begin{tabular}{p{0.2\textwidth}p{0.8\textwidth}}
\textbf{Name} & \textbf{Description} \\ \hline
Hidden size & Size of the hidden layers \\
Hidden layers & Number of hidden layers after the concatenation of the dense layers from Input Layer 1 and Input Layer 2. \\
Context size & Size of the hidden FC layer after Input Layer 1 \\
Batch size & Batch size for the stochastic gradient descent \\
Dropout & Unique dropout rate determining the proportion of neurons randomly set to zero for each output of the FC layers \\
Learning rate & Learning rate for the stochastic gradient descent. \\
Historical lags & Number of days prior to the date predicted considered for the control units outcomes. \\
Epochs Number & Number of epochs (iterations over the training dataset) required to train the model\\
\end{tabular}
\caption{Description of the hyper-parameters for the FFNN architecture.}
\label{table:hyperparams}
\end{table}

One of the challenges of our problem is that the data have an important temporal aspect. While this is not a time series problem, for a test set period, we train the model with the last observed data, making the validation step for selecting hyper-parameters difficult. To overcome this challenge, we split chronologically the pre-treatment periods in two parts: $\mathcal{T}_{\text{train}}$ and $\mathcal{T}_{\text{valid}}$. We train the model on  $\mathcal{T}_{\text{train}}$ with the backpropagation algorithm using Early Stopping, a form of regularization to avoid overfitting that consists in stopping the training when the error on the validation set increases. 
We select $\Theta$ on $\mathcal{T}_{\text{valid}}$ and store $\hat{e}$, the number of epochs it took to train the model. As a final step, we train the model with hyper-parameters $\Theta$  for $\hat{e}$ epochs on $\mathcal{T}_{\text{train}}$ and $\mathcal{T}_{\text{valid}}$, which gives an estimate $\hat{\theta}^{\text{ffnn}}$. Then, we compute the counterfactual predictions as $\hat{\mathbf{Y}}_{t}^{\text{t}}(0) = \hat{g}(\mathbf{Y}^{\text{obs,c}}, \mathbf{X}, \hat{\theta}^{\text{ffnn}})$ for $t = T_0+1, \ldots, T$.

\subsubsection{Training Details}
\label{subsection:training_details}

We present here some modeling and training tricks we used to achieve the best performance with the FFNN. 

\paragraph{Data Augmentation} Data augmentation is a well-known process to improve performances of neural networks and prevent overfitting. It is often used for computer vision tasks such as image classification \citep{shorten2019survey}. It consists in augmenting the dataset by performing simple operations such as rotation, translation, symmetry, etc. We perform one type of data augmentation, the homothety, which consists in increasing or reducing the (inputs, outputs) pair. We decompose it into the following steps. Let $a$ denote the homothety maximum coefficient, typically an integer between 1 and 4. 
For each batch in the stochastic gradient descent algorithm, we multiply each sample, inputs and outputs, by a random number uniformly distributed between $1/a$ and $a$.

\paragraph{Ensemble Learning} The ensemble learning algorithm relies on the intuition that the average performance of good models can be better than the performance of a single best model \citep{sagi2018ensemble}. We take a specific case of ensemble learning, where we consider as ensemble the 15 best models that provide the lowest MSE on the validation set from the hyper-parameter search. For each model $k=1,\ldots,15$, we store the set of hyper-parameters $\Theta_k$ and the number of training epochs $\hat{e}_k$. We train each model on the pre-treatment period to estimate $\hat{\theta}_k^{\text{ffnn}}$. We compute the counterfactuals  $\hat{\mathbf{Y}}_{t}^{\text{t} k}(0) = \hat{g}_k(\mathbf{Y}^{\text{obs,c}}, \mathbf{X}, \hat{\theta}_k^{\text{ffnn}})$ and the predicted outcome is $\hat{\mathbf{Y}}_{t}^{\text{t}}(0) = \frac{1}{15} \sum_{k=1}^{15} \hat{\mathbf{Y}}_{t}^{\text{t} k}(0)$ for $t = T_0+1, \ldots, T$.

\section{Application}
\label{section:experimental_setting}

This work was part of a large project with a major North American airline, Air Canada, operating a worldwide network. The objective of the overall project was to improve the accuracy of the demand forecasts of multiple ODs in the network. In this work, the new demand forecasting algorithm acts as the treatment. The details about the treatment is not part of this paper but it drove some of the decisions, especially regarding the selection of the treated and control units. The units correspond to the different ODs in the network and the outcome of interest is the revenue. In this paper, we present a computational study of a simulated treatment effect (ground truth impact is known). This was part of the validation work done prior to the PoC. Due to the uncertainty regarding the required duration of the treatment period, we planned for a period of 6 months in our validation study. For the sake of completeness, we also analyze the results for shorter treatment periods. 
Unfortunately, the Covid-19 situation hit the airline industry during the time of the PoC. It drastically changed the revenue and the operated flights making it impossible to
assess the impact of the demand forecasts.

In the next section, we first provide details of our experimental setting. Next, in Section~\ref{subsection:perd_perf}, we present the prediction performances of the models. In Section~\ref{subsection:simulate_treatment}, we report results from a simulation study designed to estimate the revenue impact.

\subsection{Experimental Setup and Data}
\label{subsection:application_details}

\paragraph{Treatment Effect Definition} 
There are two ways of considering the daily revenue yielded from bookings: by \textit{flight date} or by \textit{booking issue date}. The former is the total revenue at day $t$ from bookings for flights departing at $t$, while the latter is the total revenue at day $t$ from bookings made at $t$, for all possible departure dates for the flight booked. For our study, we consider the issue date revenue as it allows for a better estimation of the treatment effect. Indeed, as soon as the treatment starts at day $T_0+1$, all bookings are affected and thus the issue date revenue is affected. Hence, $Y_{it}(0)$ designates the untreated issue date revenue of OD $i$ at day $t$. The treatment period is 6 months, i.e., $T_1=181$ days. 
The drawback of the flight date revenue is that only a subset of the flights is completely affected by the treatment, hence leading to an underestimation of the treatment effect. Only flights whose booking period starts at $T_0+1$ (or after) and for which the treatment period lasts for the full duration of the booking period, approximately a year, are completely affected.

\paragraph{Selection of Treated Units} The selection of the treated ODs was the result of discussions with the airline. The objective was to have a sample of ODs representative of the North-American market, while satisfying constraints related to the demand managers in charge of those ODs. 
We select 15 non-directional treated ODs, i.e., 30 directional treated ODs ($N^{\text{t}}=30$). For instance, if Montreal-Boston was treated, then Boston-Montreal would be treated as well. The selected 30 ODs represent approximately 7\% of the airline's yearly revenue.

\paragraph{Selection of Control Units} The selection of control units depends on the treated units. Indeed, a change of the demand forecasts for an OD affects the RMS which defines the booking limits. Due to the network effect and the potential leg-sharing among ODs, this would in turn affect the demand for other ODs. With the objective to select control units that are \emph{unaffected} by the treatment, we use the following restrictions: 
\begin{itemize}
    \item Geographic rule: for each treated OD, we consider two perimeters centered around the origin and the destination airports,  respectively. We exclude all other OD pairs where either the origin or the destination is in one of the perimeters. 
    \item Revenue ratio rule: for all ODs operated by the airline in the network, different from the treated ODs, we discard the ones where at least 5\% of the itineraries have a leg identical to one of the treated ODs. This is because new pricing of OD pairs can affect the pricing of related itineraries, which in turn affects the demand. 
    \item Sparse OD rule: we exclude seasonal ODs, i.e., those that operate only at certain times of the year. Moreover, we exclude all OD pairs that have no revenue on more than 85\% of points in our dataset.   
\end{itemize}

From the remaining set of ODs, we select the 40 most correlated ODs for each treated OD. The correlation is estimated with the Pearson correlation coefficient. These rules led to $N^{\text{c}} = 317$ control units. We note that this selection is somewhat different from the literature, due to the network aspect of the airline operations and the abundance of potential control units. In \cite{abadie2010}, for instance, only a few controls are selected based on two conditions: (i) they have similar characteristics as the treated units and (ii) they are not affected by the treatment. The geographic restriction and the revenue ratio rule correspond to condition (ii). The sparse OD rule allows to partially ensure condition (i) as the treated ODs are frequent ODs from the airline's network. Considering a large number of controls has the advantage to potentially leverage the ability of deep learning models to capture the relevant information from a large set of features. 

We ran several experiments with a larger set of control units, given that the geographic rule, the revenue ratio rule and the sparse OD rule were respected. In the following, we report results for the set of controls described above, as they provided the best performance.

\paragraph{Models and Estimators}
We compare the performance of the models and estimators detailed in Section~\ref{section:existing_methods}:
\begin{itemize}
    \item DID: Difference-in-Differences
    \item SC: Abadie-Diamond-Hainmueller Synthetic Controls 
    \item CR-EN: Constrained Regressions with elastic-net regularization
    \item CR: CR-EN model with $\lambda^{\text{CR}} = 0$ and $\alpha^{\text{CR}}=0$
    \item RSC: Robust Synthetic Controls 
    \item MCNNM: Matrix Completion with Nuclear Norm Minimization
    \item FFNN: Feed-Forward Neural Network with Ensemble Learning. The external features of the FFNN are the day of the week and the week of the year. We compute a circular encoding of these two features using their polar coordinates to ensure that days 0 and 1 (respectively, week 52 and week 1 of the next year) are as distant as days 6 and days 0 (respectively, week 1 and week 2). 
\end{itemize}
We started the analysis by investigating the approach often used in practice, which consists in comparing the year-over-year revenue. The counterfactual revenue is the revenue obtained in the same period of the previous year. We ruled out this approach due to its poor performance, both in terms of accuracy and variance. We provide details in Section~\ref{sec:labTotRevPred}, where we discuss the results.

\paragraph{Data} The observed untreated daily issue date revenue covers the period from January 2013 to February 2020 for all control and treated units. This represents 907,405 data points. To test the performances of the different models, we select random periods of 6 months and predict the revenue values of the 30 treated ODs. In the literature, most studies use a random assignment of the pseudo-treated unit instead of a random assignment of treated periods. In our application, switching control units to treated units is challenging as the control set is specific to the treated units. Hence our choice of random assignment of periods. We refer to those periods as \textit{pseudo-treated} as we are interested in retrieving the observed values. To overcome the challenges described in Section~\ref{subsection:seq_valid}, we select random periods late in the dataset, between November 2018 and February 2020.

\paragraph{Two scenarios for the target variables.} We consider two scenarios for the target variables: In the first -- referred to as $S1$ -- we aggregate the 30 treated units to a single one. In the second -- referred to as $S2$ -- we predict the counterfactual revenue for each treated unit separately. For both scenarios, our interest concerns the total revenue $Y_t = \sum_{i \in N^{\text{t}}} Y_{it}$. In the following, we provide more details.

In $S1$, we aggregate the outcomes of the treated units to form one treated unit, even though the treatment is applied to each unit individually. The missing outcomes, i.e., the new target variables, are the values of $Y_t^{\text{agg}}$, where 
\begin{equation}
   \textit{(S1)} \quad Y_t^{\text{agg}} = \sum_{i = 1}^{N^{\text{t}}} Y_{i t}. 
\end{equation}
The models DID, SC, CR, CR-EN are in fact regressions on $Y_t^{\text{agg}}$ with control unit outcomes as variates. For the models RSC and MCNNM, we replace in the observation matrix $\mathbf{Y}^{\text{obs}}$ the $N^{\text{t}}$ rows of the treated units revenue with the values of $Y_t^{\text{agg}}$, for $t = 1, \ldots, T$. All models estimate $\hat{Y}_t^{\text{agg}}$, for $t = 1, \ldots, T$, and $\hat{Y}_t = \hat{Y}_t^{\text{agg}}$.

In $S2$, we predict the counterfactual revenue for each treated OD. For models SC, DID, CR, CR-EN, MCNNM and RSC, this amounts to considering each treated unit as independent from the others and we estimate a model on each treated unit. For FFNN, we relax the independence assumption so that the model can learn the dependencies and predict the revenue for each treated unit simultaneously. We have an estimate of the revenue for each pair (unit, day) in the pseudo-treatment period. Then, we estimate the total revenue at each period as the sum over each estimated pair, namely
\begin{equation}
 \textit{(S2)} \quad \hat{Y}_t = \sum_{i \in N^{\text{t}}} \hat{Y}_{it}.   
\end{equation}

\paragraph{Performance metrics} We assess performance by analyzing standard Absolute Percentage Error (APE) and Root Mean Squared Error (RMSE). In addition, the bias of the counterfactual prediction model is an important metric as it, in turn, leads to a biased estimate of the impact. In our application, the observable outcome is the issue date net revenue from the bookings whose magnitude over a 6-month treatment period is measured in millions. A pseudo-period $p$ has a length $T_{1p}$ and we report for each $p$ the percentage estimate of the total error
\begin{equation}
    \text{tPE}_p = \frac{\sum_{t=1}^{T_{1p}}  \hat{Y}_{t} - \sum_{t=1}^{T_{1p}}  Y_{ t}}{ \sum_{t=1}^{T_{1p}}  Y_{ t}} \times 100.
    \label{eq:bias}
\end{equation}
This metric allows us to have insights on whether the model tends to overestimate or underestimate the total revenue, which will be at use when estimating the revenue impact. We also report $\text{tAPE}_p$, the absolute values of $\text{tPE}_p$ for a period $p$
\begin{equation}
    \text{tAPE}_p = \frac{|\sum_{t=1}^{T_{1p}}  \hat{Y}_{t} - \sum_{t=1}^{T_{1p}}  Y_{ t}|}{ \sum_{t=1}^{T_{1p}}  Y_{ t}} \times 100.
    \label{eq:tape}
\end{equation}

We present the results of $S1$ and $S2$ in the following. For confidentiality reasons, we only report relative numbers in the remainder of the paper with the focus of comparing the different models.

\subsection{Prediction Performance}
\label{subsection:perd_perf}

In this section, we start by analyzing the performance related to predicting daily revenue, followed by an analysis of total predicted revenue in Section~\ref{sec:labTotRevPred}.

\subsubsection{Daily Predicted Revenue}

We assess the performances of the models at each day $t$ of a pseudo-treatment period, i.e., the prediction error on $\hat{Y}_t$ at each day $t$. We compute the errors for each $t$ and report the values average over all the pseudo-treatment period $p$, namely
\begin{equation}
    \text{MAPE}_p = \frac{1}{T_1} \sum_{t=1}^{T_1}\frac{|\hat{Y}_t - Y_t|}{Y_t}, \quad \text{RMSE}_p = \sqrt{\frac{1}{T_1}\sum_{t=1}^{T_1}(\hat{Y}_t - Y_t)^2}.
\end{equation}
For confidentiality reasons, we report a scaled version of $\text{RMSE}_p$ for each $p$, which we refer to as $\text{RMSE}_p^\text{s}$. We use the average daily revenue of the first year of data as a scaling factor.

\begin{figure}[!htbp]
    \begin{subfigure}[b]{\linewidth}
        \includegraphics[width=\linewidth]{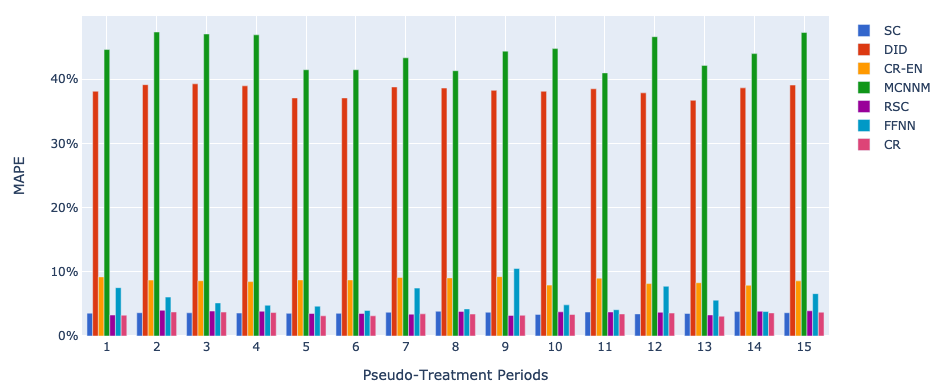}
        \caption{$S1$ (one model for a single aggregated unit)}
        \label{fig:mape_daily_s1}  
    \end{subfigure}
    \begin{subfigure}[b]{\linewidth}
        \includegraphics[width=\linewidth]{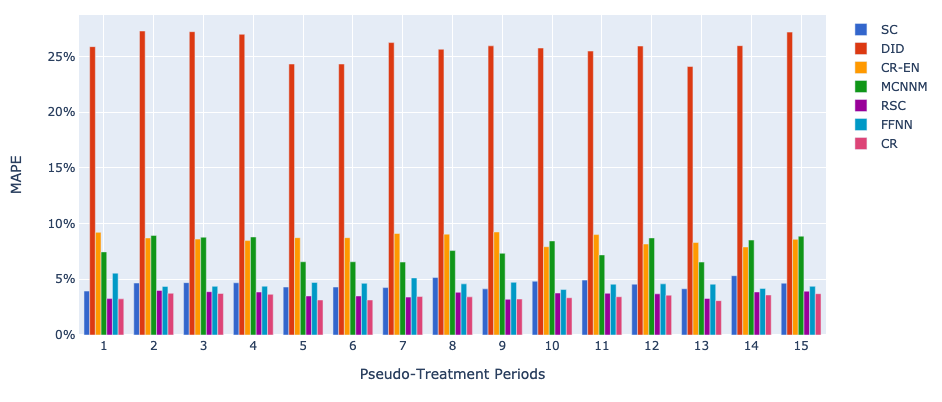}
        \caption{$S2$ (one model per treated unit)}
        \label{fig:mape_daily_s2}
    \end{subfigure}
\caption{Values of daily error, $\text{MAPE}_p$, in each pseudo-treatment period (note that the y-axis has a different scale in the two graphs).}
\label{fig:mape_daily}
\end{figure}

\begin{figure}[!htbp]
    \begin{subfigure}[b]{\linewidth}
        \includegraphics[width=\linewidth]{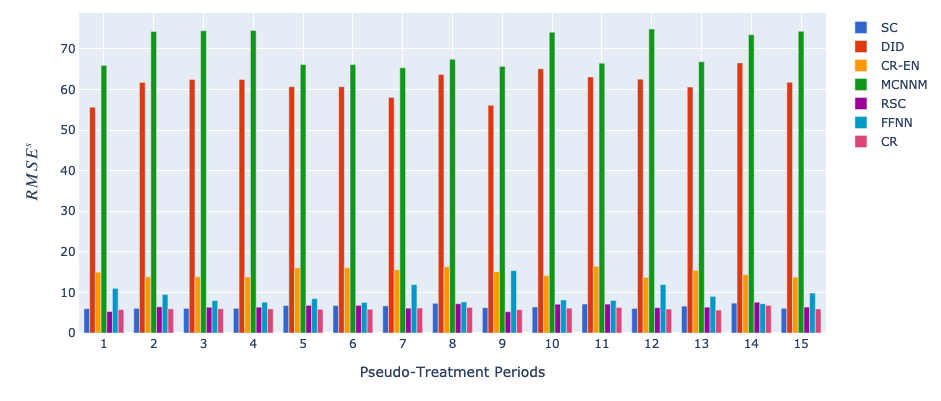}
        \caption{$S1$ (one model for a single aggregated unit)}
        \label{fig:rmse_daily_s1}  
    \end{subfigure}
    \begin{subfigure}[b]{\linewidth}
        \includegraphics[width=\linewidth]{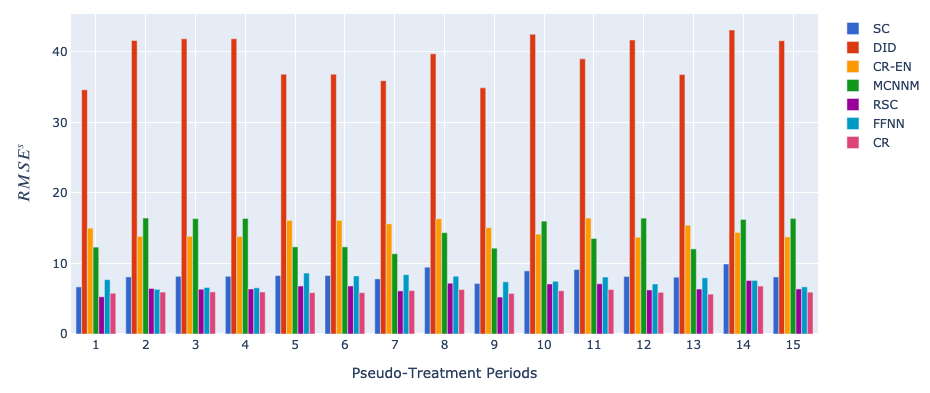}
        \caption{$S2$ (one model per treated unit)}
        \label{fig:rmse_daily_s2}
    \end{subfigure}
\caption{Values of daily $\text{RMSE}^\text{s}$ in each pseudo-treatment period (note that the y-axis has a different scale in the two graphs).}
\label{fig:rmse_daily}
\end{figure}

Figures~\ref{fig:mape_daily} and \ref{fig:rmse_daily} present $\text{MAPE}_p$ and $\text{RMSE}_p^\text{s}$ for $p=1,\ldots,15$, where the upper graph of each figure shows results for $S1$ and the lower the results for $S2$, respectively. We note that the performance is stable across pseudo-treated periods for all models. The values of $\text{MAPE}_p$ at each period $p$ of SC, RSC and CR models are below 5\%  while for FFNN it is only the case in $S2$. This is important, as the impact we wish to measure is less than this order of magnitude.

Table~\ref{table:daily} reports the values of the metrics averaged over all pseudo-treatment periods for settings $S1$ and $S2$, i.e., $\text{MAPE} = \frac{1}{15} \sum_{p=1}^{15} \text{MAPE}_p$ and $\text{RMSE}^\text{s} = \frac{1}{15} \sum_{p=1}^{15} \text{RMSE}_p^\text{s}$. The results show that the best performance for both metrics and in both scenarios is achieved by CR model. On average, it reaches a MAPE of 3.4\% and $\text{RMSE}^{\text{s}}$ of 6.0. It achieves better results than CR-EN model. This is because we have $T \gg N$ and there are hence enough data to estimate the coefficients without regularization. 

\begin{table}[!htbp]
\centering
\begin{tabular}{lll|ll}
 & \multicolumn{2}{c}{$S1$} & \multicolumn{2}{c}{$S2$} \\
\multicolumn{1}{l|}{} & MAPE & $\text{RMSE}^\text{s}$ & MAPE & $\text{RMSE}^\text{s}$ \\ \hline
\multicolumn{1}{l|}{CR} & \textbf{3.4}\% & \textbf{6.0}  & \textbf{3.4\%}  &  \textbf{6.0}\\
\multicolumn{1}{l|}{CR-EN} & 8.6\% & 15.0  & 8.6\% & 15.0   \\
\multicolumn{1}{l|}{DID} & 38.3\% & 61.4 & 25.9\% & 39.2 \\
\multicolumn{1}{l|}{FFNN} & 5.8\% & 9.4  & 4.6\% & 7.5 \\
\multicolumn{1}{l|}{MCNNM} & 44.2\% & 70.0 & 7.8\% & 14.3 \\
\multicolumn{1}{l|}{RSC} & 3.6\% & 6.5 & 3.6\%  & 6.5 \\
\multicolumn{1}{l|}{SC} & 3.6\% & 6.5 & 4.6\% & 8.3
\end{tabular}
\caption{Average of the daily MAPE and $\text{RMSE}^\text{s}$ over all pseudo-treatment periods.}
\label{table:daily}
\end{table}

Models DID and MCNNM have poor performance in $S1$. This is due to the difference in magnitude between the treated unit and the control units.  In $S2$, the performance is improved because we build one model per treated unit. Each treated unit is then closer to the controls in terms of magnitude. Due to the constraint \eqref{eq:cst_weights} of equal weights, DID model is not flexible enough and its performance does not reach that of the other models.

The FFNN model improves the MAPE by 1.2 points from $S1$ to $S2$. The neural network models the dependencies between the treated ODs and gain accuracy by estimating the revenue of each treated OD.

The advantage of $S2$ is that we predict separately the outcome for each unit at each day. In addition to computing the error between $\hat{Y}_t$ and $Y_t$ for each pseudo-treatment period, we can also compute the error between $\hat{Y}_{it}$ and $Y_{it}$, for $i=1,\ldots,N^{\text{t}}$, and $t=1,\ldots,T_1$, namely
\begin{equation}
    \text{MAPE}^{\text{od}}_i = \frac{1}{T_1} \sum_{t=1}^{T_1}\frac{|\hat{Y}_{it} - Y_{it}|}{Y_{it}}, \quad \text{MAPE}^{\text{od}} = \frac{1}{N^{\text{t}}} \sum_{i=1}^{N^\text{t}} \text{MAPE}^{\text{od}}_i.
\end{equation}

\begin{figure}[!htbp]
  \includegraphics[width=1.1\linewidth]{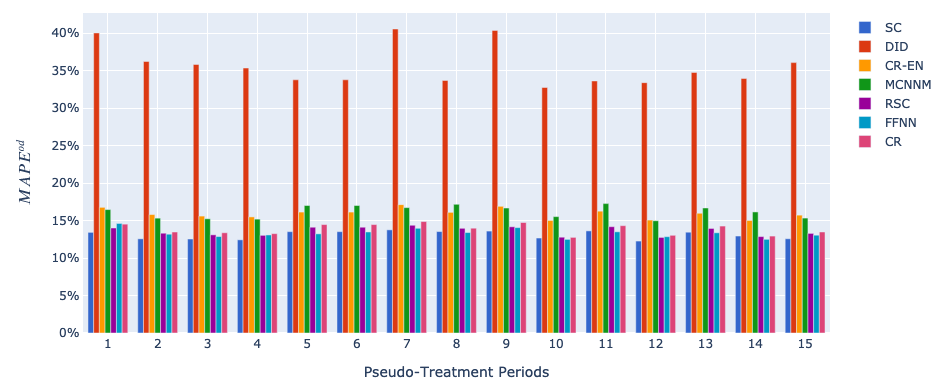}
  \caption{$\text{MAPE}^{\text{od}}$ for each pseudo-treatment period in $S2$.}
  \label{fig:mape_disagg_od}
\end{figure}

Figure~\ref{fig:mape_disagg_od} presents the values of $\text{MAPE}^{\text{od}}$ for each pseudo-treatment period, and 
Table~\ref{table:mape_disagg_od} reports the average value of $\text{MAPE}^{\text{od}}$ over all pseudo-treatment periods. It shows that results are consistent across periods. Method SC reaches the best accuracy, with on average 13.1\% of error for the daily revenue of one treated OD. The FFNN model has a similar performance with 13.3\% of error on average. We conclude that estimating the counterfactual revenue of one OD is difficult and we gain significant accuracy by aggregating over the treated ODs. In the remainder of the paper, we only consider models CR, CR-EN, FFNN, RSC and SC as they perform best. 

\begin{table}[!htbp]
\centering
\begin{tabular}{l|c}
 	    & $\text{MAPE}^{\text{od}}$ 	         \\ \hline
CR 	    & 13.8\%	   \\
CR-EN 	& 16.0\%	    \\
DID 	& 35.6\%	    \\
FFNN 	& 13.3\%	    \\
MCNNM 	& 16.2\%	   \\
RSC 	& 13.6\%	    \\
SC 	    & $\mathbf{13.1}$\%
\end{tabular}
\caption{$\text{MAPE}^{\text{od}}$ averaged over all pseudo-treatment periods in $S2$.}
\label{table:mape_disagg_od}
\end{table}

\subsubsection{Total Predicted Revenue} \label{sec:labTotRevPred}

In this section, we analyze the models' performance over a complete pseudo-treatment period. We first consider a pseudo-treatment period of 6 months, and we then analyze the effect of a reduced length.

\begin{figure}[!htbp]
    \begin{subfigure}[b]{\linewidth}
        \includegraphics[width=\linewidth]{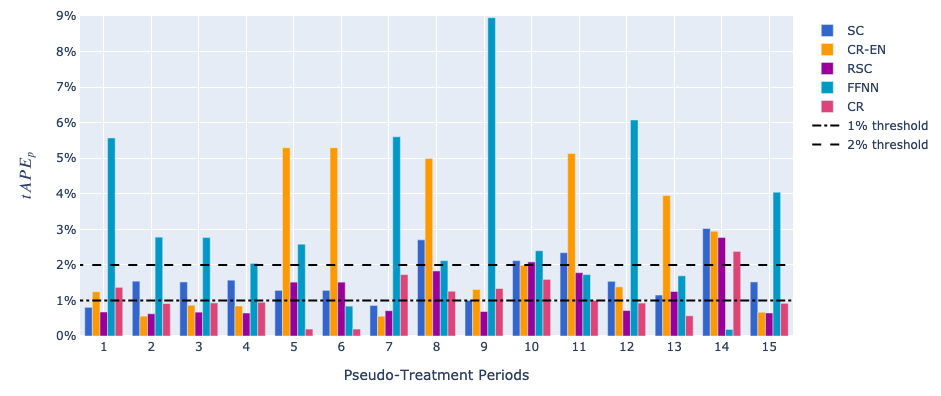}
        \caption{$S1$ (one model for a single aggregated unit)}
        \label{fig:mape_agg_6months_s1}  
    \end{subfigure}
    \begin{subfigure}[b]{\linewidth}
        \includegraphics[width=\linewidth]{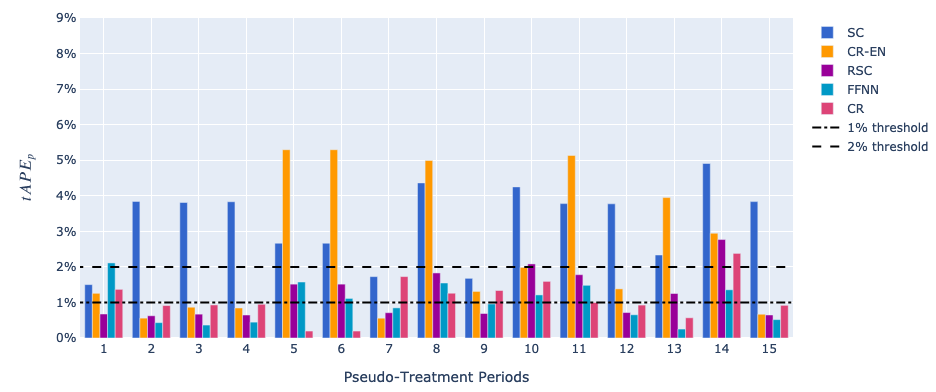}
        \caption{$S2$ (one model per treated unit)}
        \label{fig:mape_agg_6months_s2}
    \end{subfigure}
\caption{Values of $\text{tAPE}_p$ for each pseudo-treatment period.}
\label{fig:mape_agg_6months}
\end{figure}

Figure~\ref{fig:mape_agg_6months} presents the value of $\text{tAPE}_p$ defined in~\eqref{eq:tape} for pseudo-treatment periods $p=1,\ldots,15$. The upper graph shows the results for $S1$ and the lower the results for $S2$, respectively. To illustrate treatment impacts' order of magnitude, we depict the 1\% and 2\% thresholds in dashed lines.
We note that FFNN and CR-EN models have higher variance than SC, CR and RSC methods which stay below 3\% of error at each period. Moreover, the model FFNN is stable across all periods for $S2$.

Table~\ref{table:6months} reports the values of $\text{tAPE} = \frac{1}{15} \sum_{p=1}^{15} \text{tAPE}_p$ for each model. All models are able to predict the total 6-months counterfactual revenue with less than 3.5\% of error on average, in both settings. For $S1$, the CR method reaches the best performance, with 1.1\% error on average and, for $S2$, the best is the FFNN model with 1.0\% average error. 

\begin{table}[!htbp]
\centering
\begin{tabular}{ll|l}
 & \multicolumn{1}{c|}{$S1$} & \multicolumn{1}{c}{$S2$} \\
\multicolumn{1}{l|}{} & tAPE & tAPE \\ \hline
\multicolumn{1}{l|}{CR} & \textbf{1.1}\%  & 1.1\%  \\
\multicolumn{1}{l|}{CR-EN} & 2.5\% & 2.5\% \\
\multicolumn{1}{l|}{FFNN} & 3.3\% & \textbf{1.0\%} \\
\multicolumn{1}{l|}{RSC} & 1.2\% & 1.2\% \\
\multicolumn{1}{l|}{SC} & 1.6\% & 3.3\%
\end{tabular}
\caption{tAPE over all pseudo-treatment periods}
\label{table:6months}
\end{table}

\begin{figure}[!htbp]
    \begin{subfigure}[b]{\linewidth}
        \includegraphics[width=\linewidth]{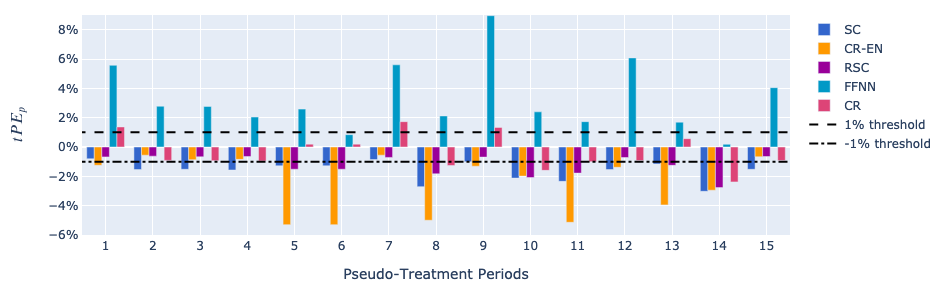}
        \caption{$S1$ (one model for a single aggregated unit)}
        \label{fig:pe_agg_6months_s1}  
    \end{subfigure}
    \begin{subfigure}[b]{\linewidth}
        \includegraphics[width=\linewidth]{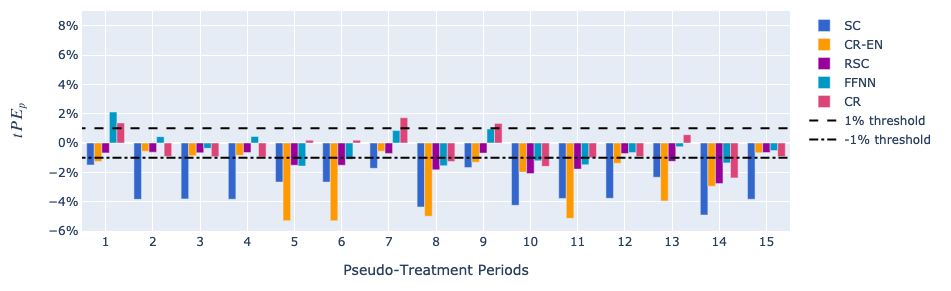}
        \caption{$S2$ (one model per treated unit)}
        \label{fig:pe_agg_6months_s2}
    \end{subfigure}
\caption{Values of $\text{tPE}_p$ for each pseudo-treatment period $p=1,\ldots,15$.}
\label{fig:pe_agg_6months}
\end{figure}

We present in Figure~\ref{fig:pe_agg_6months} the values of $\text{tPE}_p$ defined in~\eqref{eq:bias} at each period $p=1,\ldots,15$. It shows that for $S1$, the FFNN model systematically overestimates the total counterfactual revenue while SC, CR-EN and RSC methods systematically underestimate it. For $S2$, we observe the same behavior for models SC, CR-EN and RSC while both FFNN and CR methods either underestimate or overestimate the counterfactual revenue. 

\paragraph{Length of the treatment period}
We now turn our attention to analyzing the effect of the treatment duration period on performance.
For this purpose, we study the variations of $\text{tAPE}_p$ for different values of $T_1$ for the pseudo-treatment periods $p=1,\ldots,15$. We analyze the results for each period but for illustration purposes we focus only on the second one.
We report the values for all the other periods in  Appendix~\ref{appendix:length_treatment} (the general observations we describe here remain valid). 

\begin{figure}[!htbp]
    \begin{subfigure}[b]{\linewidth}
        \includegraphics[width=\linewidth]{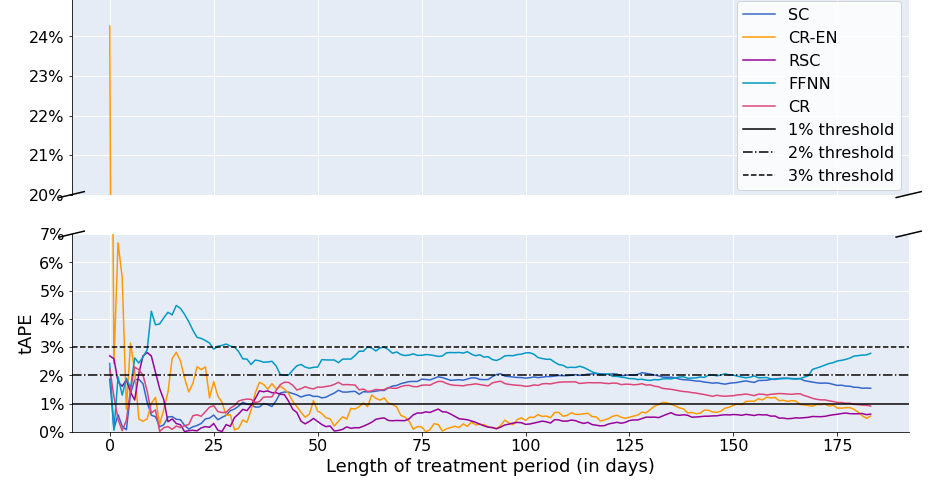}
        \caption{$S1$ (one model for a single aggregated unit)}
        \label{fig:ape_vary_T_s1}  
    \end{subfigure}
    \begin{subfigure}[b]{\linewidth}
        \includegraphics[width=\linewidth]{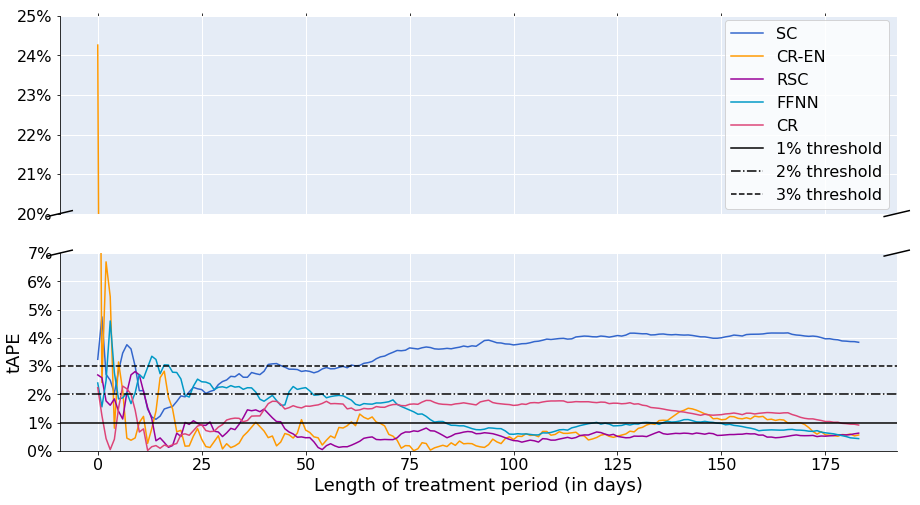}
        \caption{$S2$ (one model per treated unit)}
        \label{fig:ape_vary_T_s2}
    \end{subfigure}
\caption{Values of $\text{tAPE}_2$ varying with the length of the treatment period $T_1$.}
\label{fig:ape_vary_T}
\end{figure}

Figure~\ref{fig:ape_vary_T} presents the variations of $\text{tAPE}_2$ against the length $T_1$ for the different models. The upper graph shows the results for $S1$ and the lower one the results for $S2$, respectively. The  black lines (solid and dashed) represent the 1\%, 2\% and 3\% thresholds. In $S1$, values of $\text{tAPE}_2$ for FFNN are below 3\% from 30 days. After 30 and 39 days, respectively, $\text{tAPE}_2$ values for CR and SC are between 1\% and 2\%. Values of $\text{tAPE}_2$ are below 1\% from 68 days for CR-EN and from 43 days for RSC. In $S2$, $\text{tAPE}_2$ for FFNN is below 2\% from 52 days and below 1\% from 84 days. For CR and CR-EN, it is below 2\% from 10 days and 18 days, respectively. It is below 1\% from 44 days for RSC. Hence, the results show that the length of the treatment period can be less than six months as models are accurate after only a few weeks.

The CR, RSC and FFNN models present high accuracy with errors less than 1.2\% for the problem of counterfactual predictions on the total revenue. This is compelling since we are interested in detecting a small treatment impact. As anticipated in Section \ref{subsection:application_details}, we considered simpler approaches that are common practice. For example, comparing to year-over-year revenue. In this case, the counterfactual revenue is defined as the revenue generated during the same period but the year before. It had a poor performance, with a $\text{tAPE}$ between 7\% and 10\% at each pseudo-treatment period. This approach is therefore not accurate enough to detect small impacts. 

In the following section, we present a validation study where we simulate small impacts and assess our ability to estimate them with 
counterfactual prediction models. 

\subsection{Validation: Revenue Impact Estimate for Known Ground Truth}
\label{subsection:simulate_treatment}

We consider a pseudo-treatment period of 6 months and the setting $S2$. In this case, models FFNN, CR and RSC provide accurate estimations of the counterfactual total revenue with respectively 1\%, 1.1\% and 1.2\% of error on average over the pseudo-treatment periods. We restrict the analysis that follows to those models. We proceed in two steps:
First, we simulate a treatment by adding a noise with positive mean to the revenue of the treated units at each day of each pseudo-treatment period. We denote $\Tilde{Y}_{t}^{\text{obs}}$ the new treated value, $\Tilde{Y}_{t}^{\text{obs}} =  Y_{t}(0) \times \epsilon, \quad \epsilon \sim \text{Lognormal}(\mu_{\epsilon}, \sigma_{\epsilon}^2)$ and $\sigma_{\epsilon}^2 = 0.0005$. We simulate several treatment impacts that differ by the value of $\mu_{\epsilon}$. Second, we compute the impact estimate with~\eqref{eq:impact} from the counterfactual predictions and compare it to the actual treatment applied in the first step. We present the results for one pseudo-treatment period, $p=2$. 

Table~\ref{table:simulation_impact} reports the values of the estimated impact for different values of $\mu_{\epsilon}$. The first row shows the values for the true counterfactuals. This is used as reference, as it is the exact simulated impact. Results show that RSC and CR models overestimate the impact while FFNN model underestimates it. This is because the former underestimates the counterfactual predictions while the latter overestimates them. Due to the high accuracy of counterfactual predictions, both the underestimation and  overestimation are however small. We can detect impacts higher than the accuracy of the counterfactual prediction models. The simulation shows that we are close to the actual impact.

\begin{table}[!htbp]
\centering
\begin{tabular}{l|cccc}
Counterfactuals & $\mu_{\epsilon} = 0.01$ & $\mu_{\epsilon} = 0.02$ & $\mu_{\epsilon} = 0.03$ & $\mu_{\epsilon} = 0.05$ \\ \hline
Ground truth  & 1.0\% & 2.0\% & 3.0\% & 5.1\% \\ 
RSC   & 1.7\% & 2.6\% & 3.7\% & 5.7\% \\
CR    & 1.5\% & 2.5\% & 3.5\% & 5.6\% \\
FFNN  & 0.6\% & 1.6\% & 2.6\% & 4.7\%
\end{tabular}
\caption{Estimation of the revenue impact $\hat{\tau}$ of simulated treatment.}
\label{table:simulation_impact}
\end{table}

\begin{figure}[!htbp]
  \includegraphics[width=1.1\linewidth]{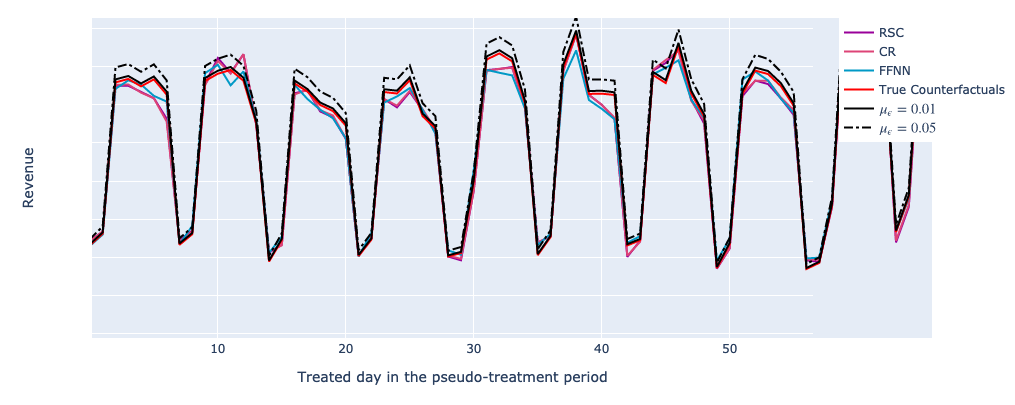}
  \caption{Daily revenue and predictions for a subset of the pseudo-treatment period 2. The labels in the y-axis are hidden for confidentiality reasons.}
  \label{fig:simulation_curves_iter14}
\end{figure}

Figure~\ref{fig:simulation_curves_iter14} presents the daily revenue on a subset of the treatment periods. The estimation of the daily revenue impact is the difference between the simulated revenue (solid and dashed black lines) and the counterfactual predictions (colored lines). This figure reveals that even though the accuracy of the daily predictions is not as good as on the complete treatment period, we can still detect even a small daily impact.

\paragraph{Prediction intervals.} It is clear that prediction intervals for the estimated revenue impact are of high importance. However, it is far from trivial to compute them
for most of the counterfactual prediction models in our setting. Under some assumptions, the CR model in setting $S1$ constitutes the exception. 
More precisely, if the residuals satisfy conditions (i) independent and identically distributed and (ii) normally distributed, then we can derive a prediction interval for the sum of the daily predicted revenue. For the simulated impacts reported in Table~\ref{table:simulation_impact}, we obtain 99\% prediction intervals with widths of 2.2\%. It means that we can detect an impact of 2\% or more with high probability.

\cite{cattaneo2019prediction} develop prediction intervals for the SC model that account for two distinct sources of randomness: the construction of the weights $\mathbf{\omega}$ and the unobservable stochastic error in the treatment period. Moreover, \cite{zhu2017deep} build prediction intervals for neural networks predictions that consider three sources of randomness: model uncertainty, model misspecification and data generation process uncertainty. Both studies focus on computing prediction intervals for \emph{each} prediction. We face an additional issue as we need a prediction interval for the sum of the predictions. As evidenced by these two studies, computing accurate prediction intervals is a challenging topic on its own and we therefore leave it for future research.

\section{Conclusion}
\label{section:conclusion}

Revenue management systems are crucial to the profitability of airlines and other industries. Due to their importance, solution providers and  airlines invest in the improvement of the different system components.
In this context, it is important to estimate the impact on an outcome such as revenue after a proof of concept. We addressed this problem using counterfactual prediction models.

In this paper, we assumed that an airline applies a treatment (a change to the system) on a set of ODs during a limited time period. We aimed to estimate the total impact over all of the treated units and over the treatment period. We proceeded in two steps. First we estimated the counterfactual predictions of the ODs' outcome, that is the outcome if no treatment were applied. Then, we estimated the impact as the difference between the observed revenue under treatment and the counterfactual predictions. 

We compared the performance of several counterfactual prediction models and a deep-learning model in two different settings. In the first one, we predicted the aggregated outcome of the treated units while in the second one, we predicted the outcome of each treated unit and aggregated the predictions. We showed that synthetic control methods and the deep-learning model reached a competitive accuracy on the counterfactual predictions, which in turn allows to accurately estimate the revenue impact. The deep-learning model reaches the lowest error of 1\% in the second setting, leveraging the dependency between treated units. The best counterfactual prediction model, which in the second setting assumes treated units are independent, reached 1.1\% of error in both settings. We showed that we can reduce the length of a treatment period and preserve this level of accuracy. This can be useful as it potentially allows to reduce the cost of proofs of concepts.

We believe that the methodology is broadly applicable to decision support systems, and not limited to revenue management (e.g., upgrade of a software, new marketing policy). It can assess the impact of a proof of concept under the following fairly mild assumptions: (i) the units under consideration (e.g., origin-destination pairs, markets, sites or products) can be divided into two subsets, one affected by the treatment and one that is unaffected (ii) time can be divided into two (not necessarily consecutive) periods, a pre-treatment period and a treatment period (iii) the outcome of interest (any objective function value, for example, revenue, cost or market share) can be measured for each unit. 

Finally, we will dedicate future research to devise prediction intervals for the sum of the counterfactual predictions, which in turn will lead to a prediction interval for the estimated impact.

\section*{Acknowledgements}
We are grateful for the invaluable support from the whole Crystal AI team who built the demand forecasting solution. 
The team included personnel from both Air Canada and IVADO Labs. In particular, we would like to thank Richard Cleaz-Savoyen and the Revenue Management team for careful reading and comments that have helped improving the manuscript. 
We also thank Florian Soudan from Ivado Labs and Pedro Garcia Fontova from Air Canada for their help and advice in training the neural network models. 
We would like to especially thank William Hamilton from IVADO Labs who has contributed with ideas and been involved in the results analysis. Maxime Cohen provided valuable comments that helped us improve the manuscript.We express our gratitude to Peter Wilson (Air Canada) who gave valuable business insights guiding the selection of control units. The project was partially funded by Scale AI. Finally, the first author would like to thank Louise Laage and the third author would like to thank Luca Nunziata and Marco Musumeci for many stimulating discussions on counterfactual prediction and synthetic control.

\section*{Appendix}

\subsection*{Length of Treatment Period}
\label{appendix:length_treatment}

We present here the results on the analysis of the length of the treatment-period for all pseudo-treatment periods. 

\begin{figure}[H]
	\begin{subfigure}[b]{0.5\linewidth}
		\includegraphics[width=\linewidth]{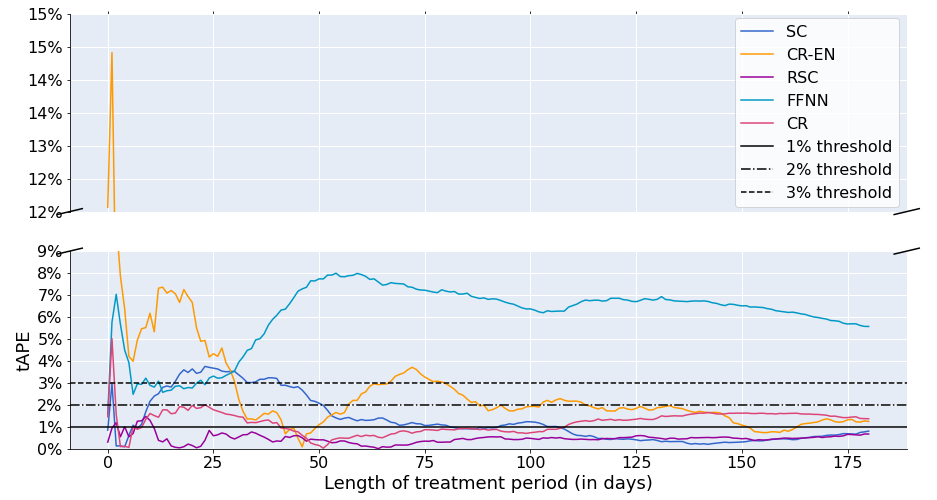}
		\caption{$S1$}
	\end{subfigure}
	\begin{subfigure}[b]{0.5\linewidth}
		\includegraphics[width=\linewidth]{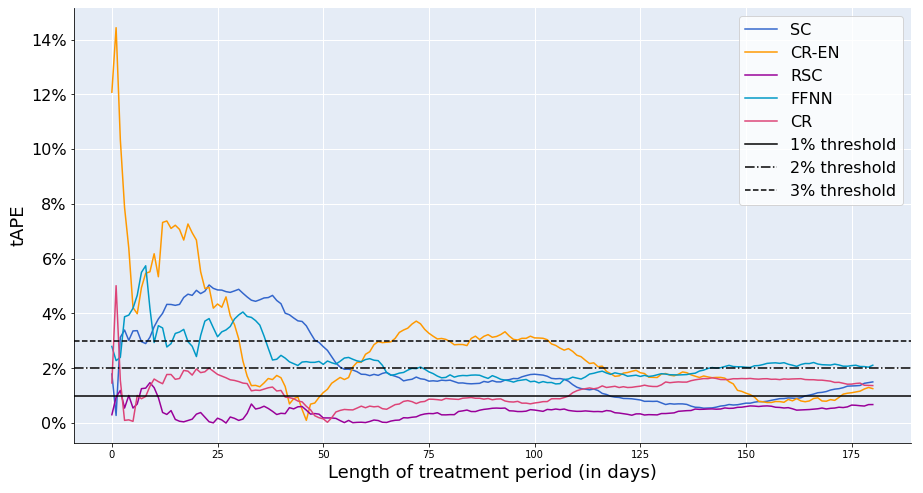}
		\caption{$S2$}
	\end{subfigure}
	\caption{Values of tAPE varying with the length of the treatment period for pseudo-treatment period 1}
\end{figure} 

\begin{figure}[H]
	\begin{subfigure}[b]{0.5\linewidth}
		\includegraphics[width=\linewidth]{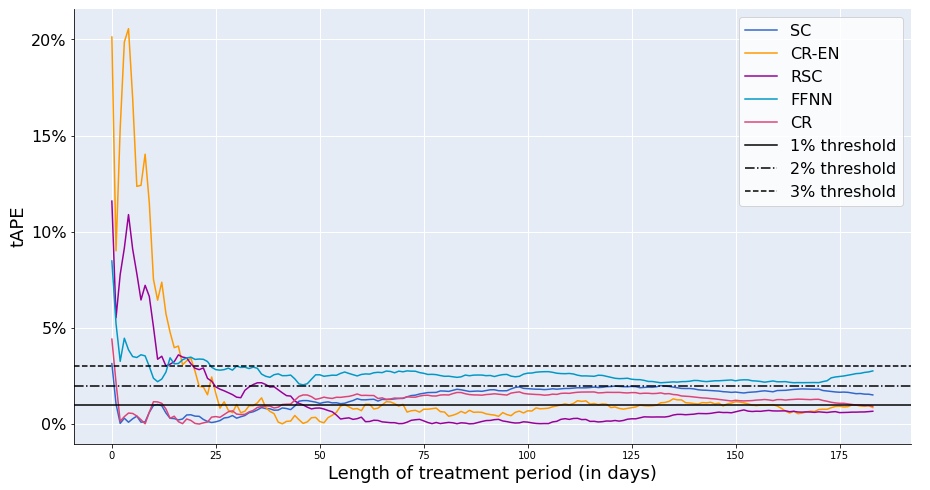}
		\caption{$S1$}
	\end{subfigure}
	\begin{subfigure}[b]{0.5\linewidth}
		\includegraphics[width=\linewidth]{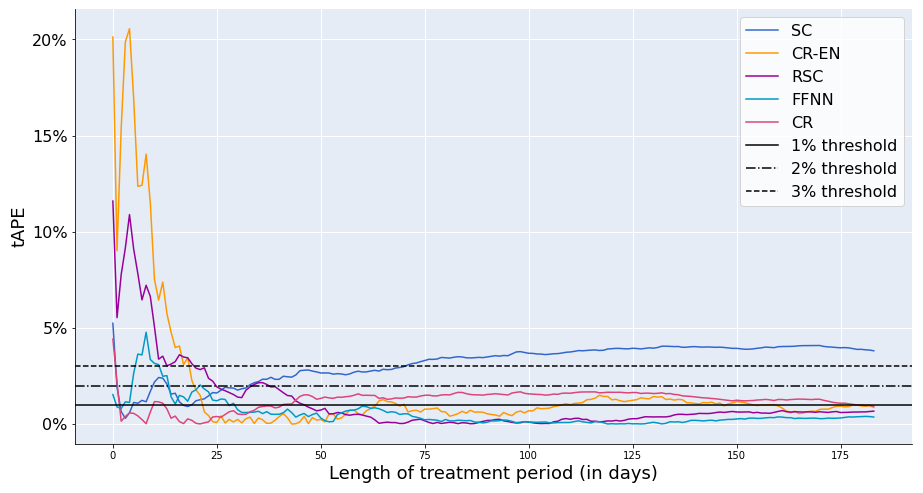}
		\caption{$S2$}
	\end{subfigure}
	\caption{Values of tAPE varying with the length of the treatment period for pseudo-treatment period 3}
\end{figure} 

\begin{figure}[H]
	\begin{subfigure}[b]{0.5\linewidth}
		\includegraphics[width=\linewidth]{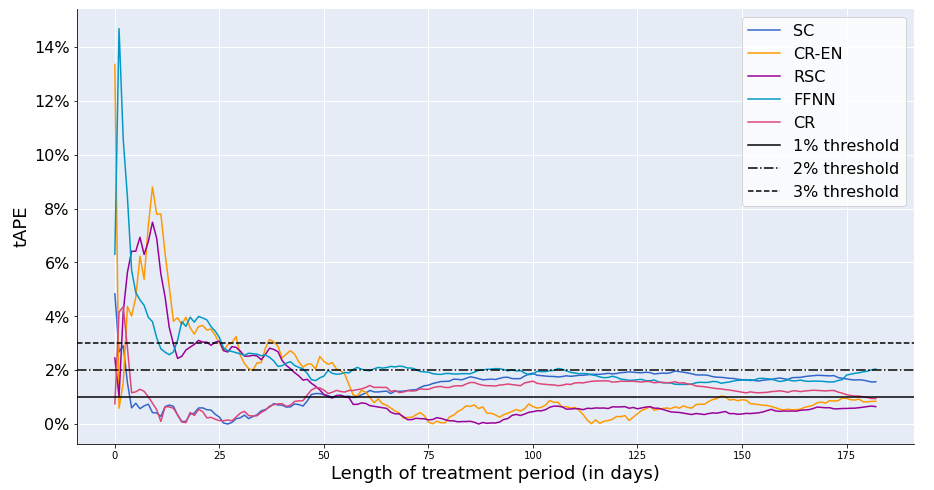}
		\caption{$S1$}
	\end{subfigure}
	\begin{subfigure}[b]{0.5\linewidth}
		\includegraphics[width=\linewidth]{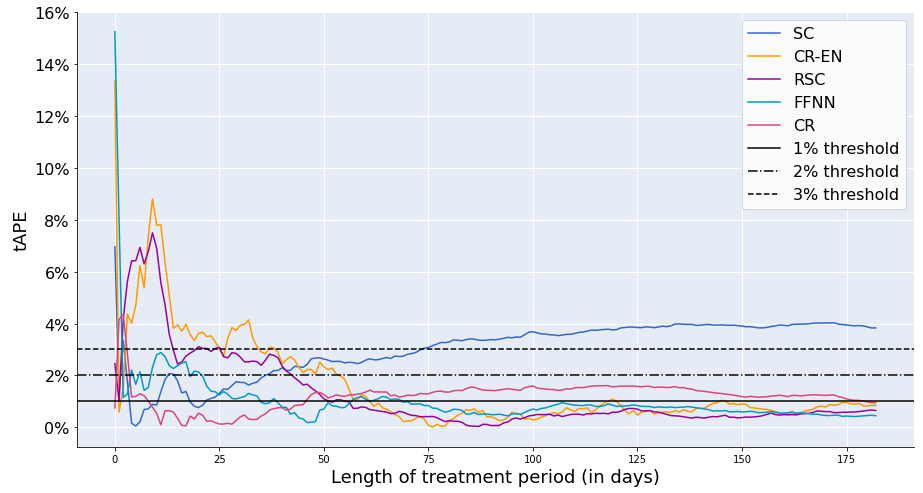}
		\caption{$S2$}
	\end{subfigure}
	\caption{Values of tAPE varying with the length of the treatment period for pseudo-treatment period 4}
\end{figure} 

\begin{figure}[H]
	\begin{subfigure}[b]{0.5\linewidth}
		\includegraphics[width=\linewidth]{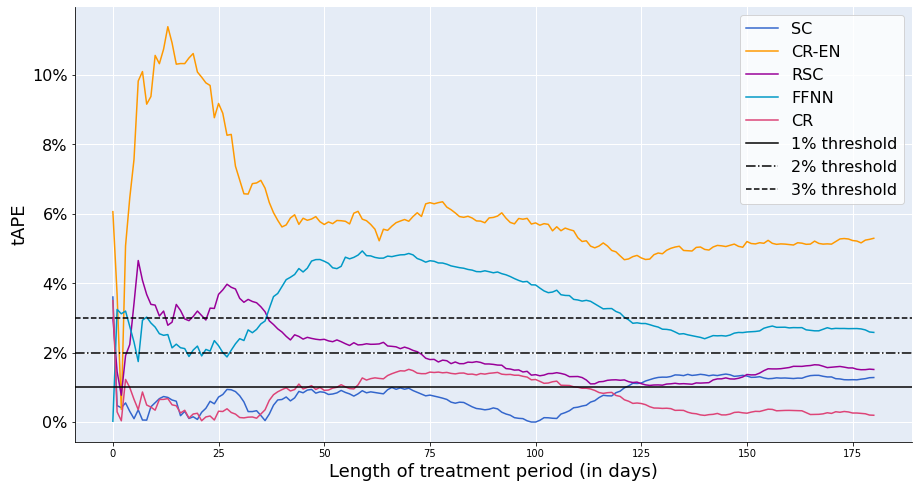}
		\caption{$S1$}
	\end{subfigure}
	\begin{subfigure}[b]{0.5\linewidth}
		\includegraphics[width=\linewidth]{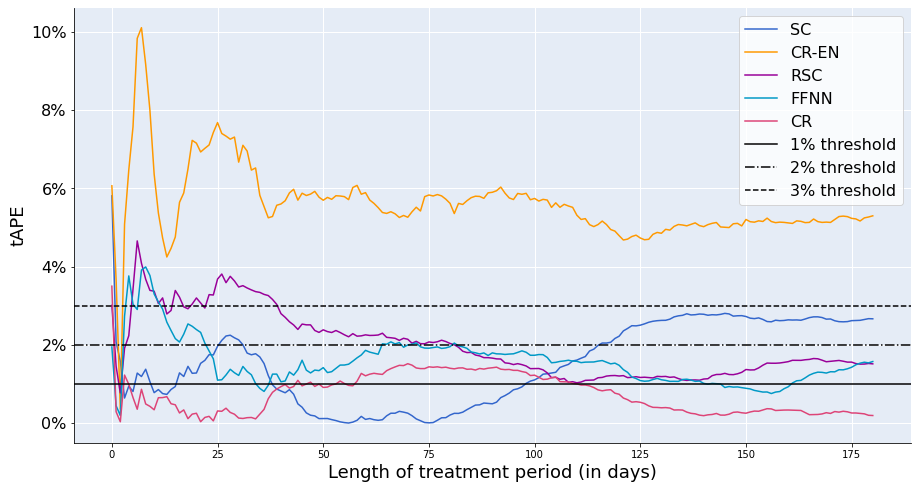}
		\caption{$S2$}
	\end{subfigure}
	\caption{Values of tAPE varying with the length of the treatment period for pseudo-treatment period 5}
\end{figure} 

\begin{figure}[H]
	\begin{subfigure}[b]{0.5\linewidth}
		\includegraphics[width=\linewidth]{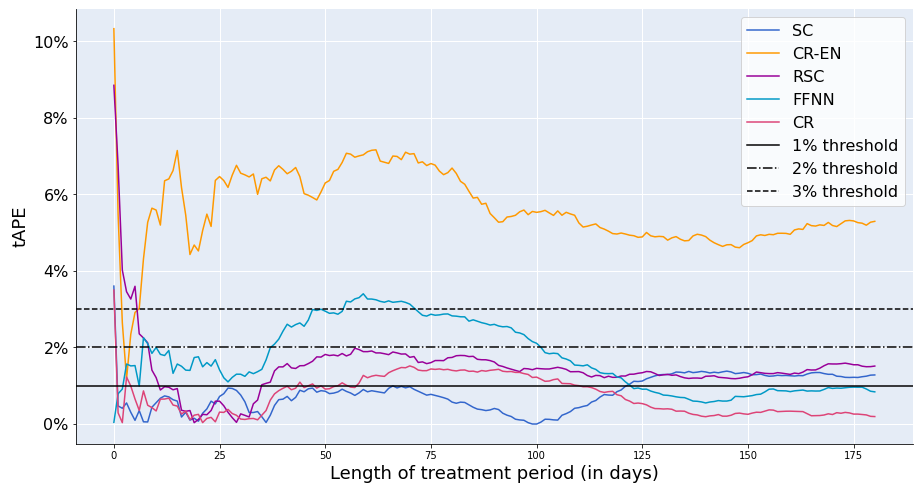}
		\caption{$S1$}
	\end{subfigure}
	\begin{subfigure}[b]{0.5\linewidth}
		\includegraphics[width=\linewidth]{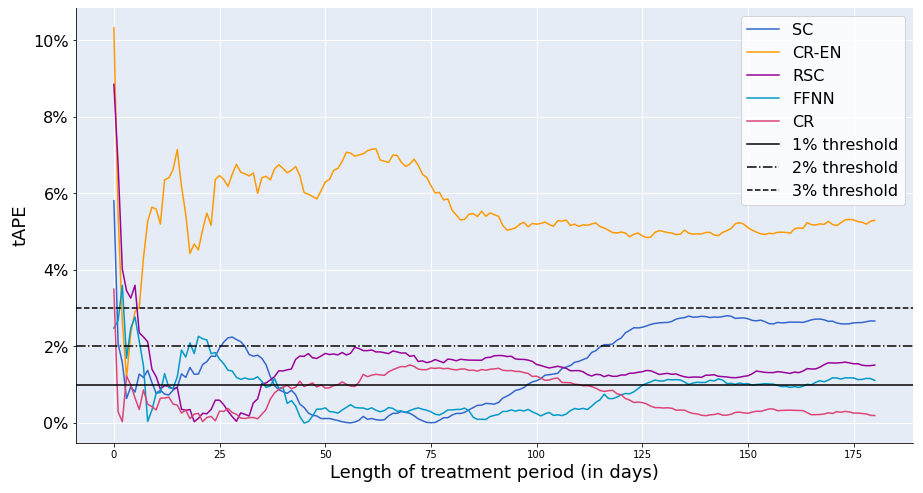}
		\caption{$S2$}
	\end{subfigure}
	\caption{Values of tAPE varying with the length of the treatment period for pseudo-treatment period 6}
\end{figure} 

\begin{figure}[H]
	\begin{subfigure}[b]{0.5\linewidth}
		\includegraphics[width=\linewidth]{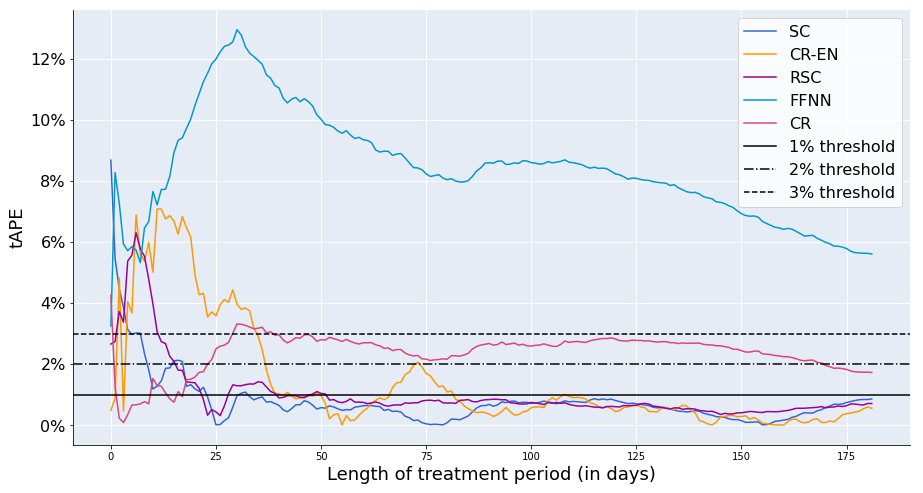}
		\caption{$S1$}
	\end{subfigure}
	\begin{subfigure}[b]{0.5\linewidth}
		\includegraphics[width=\linewidth]{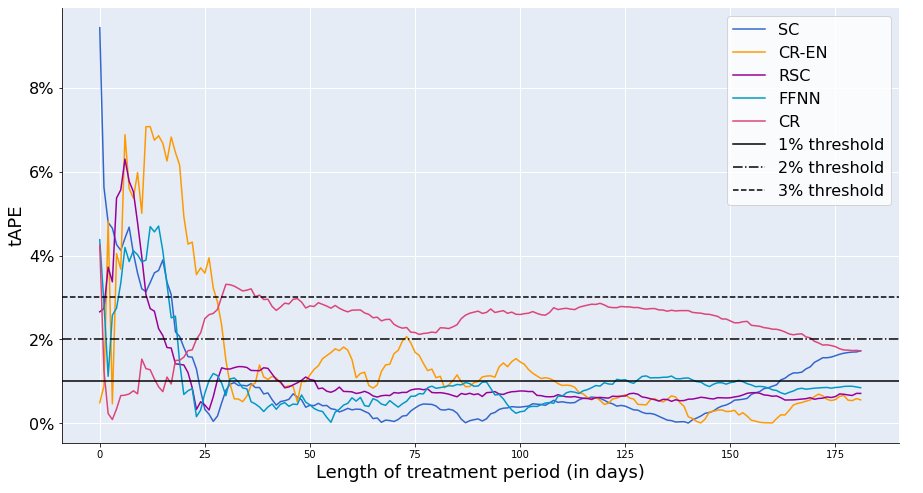}
		\caption{$S2$}
	\end{subfigure}
	\caption{Values of tAPE varying with the length of the treatment period for pseudo-treatment period 7}
\end{figure} 

\begin{figure}[H]
	\begin{subfigure}[b]{0.5\linewidth}
		\includegraphics[width=\linewidth]{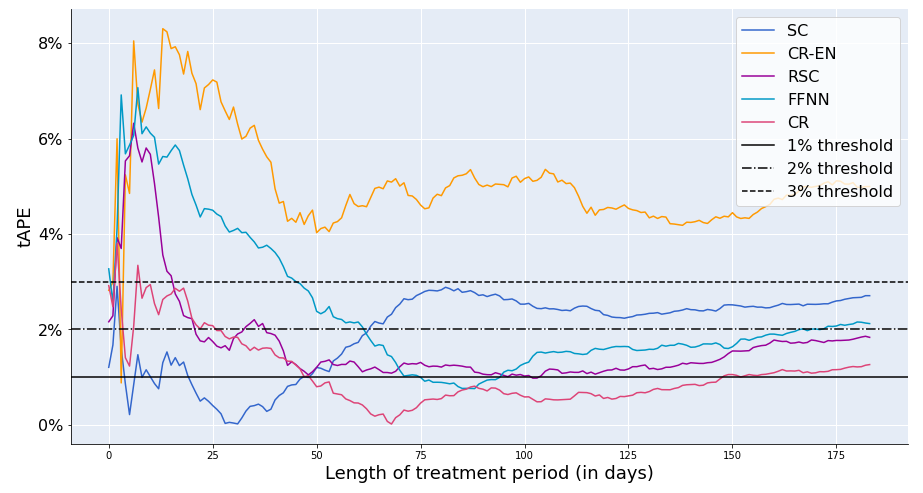}
		\caption{$S1$}
	\end{subfigure}
	\begin{subfigure}[b]{0.5\linewidth}
		\includegraphics[width=\linewidth]{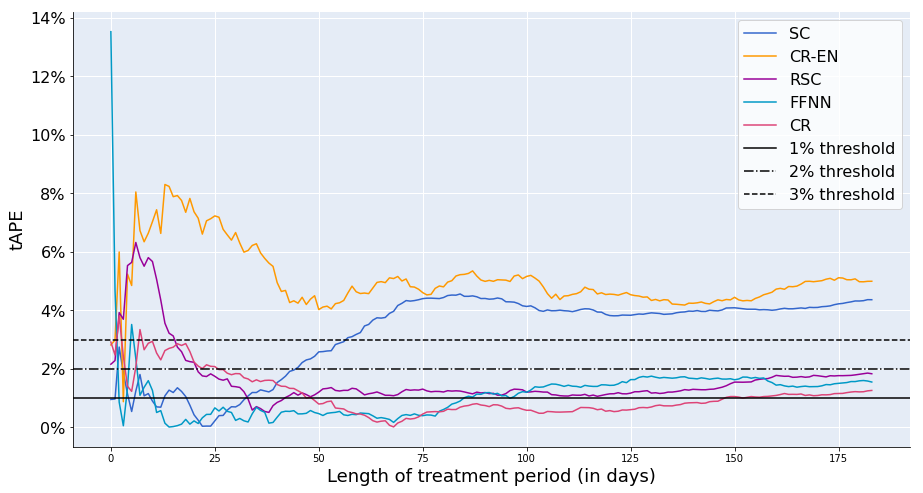}
		\caption{$S2$}
	\end{subfigure}
	\caption{Values of tAPE varying with the length of the treatment period for pseudo-treatment period 8}
\end{figure} 

\begin{figure}[H]
	\begin{subfigure}[b]{0.5\linewidth}
		\includegraphics[width=\linewidth]{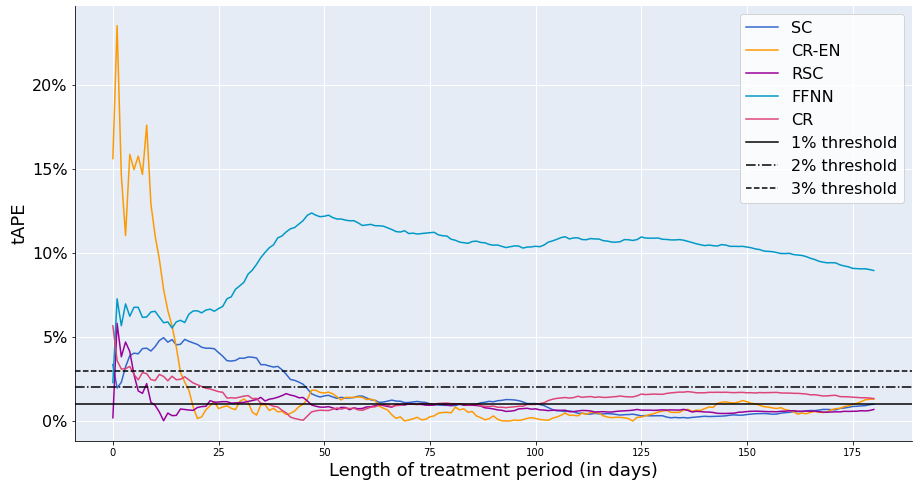}
		\caption{$S1$}
	\end{subfigure}
	\begin{subfigure}[b]{0.5\linewidth}
		\includegraphics[width=\linewidth]{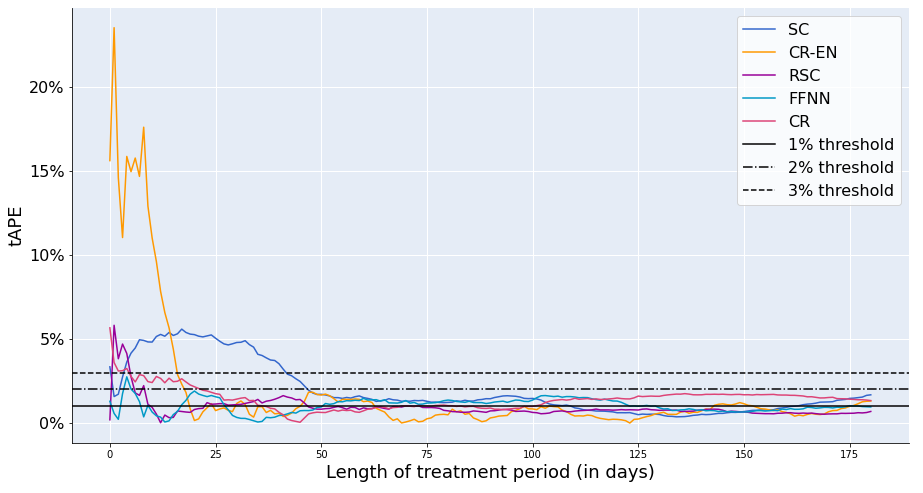}
		\caption{$S2$}
	\end{subfigure}
	\caption{Values of tAPE varying with the length of the treatment period for pseudo-treatment period 9}
\end{figure} 

\begin{figure}[H]
	\begin{subfigure}[b]{0.5\linewidth}
		\includegraphics[width=\linewidth]{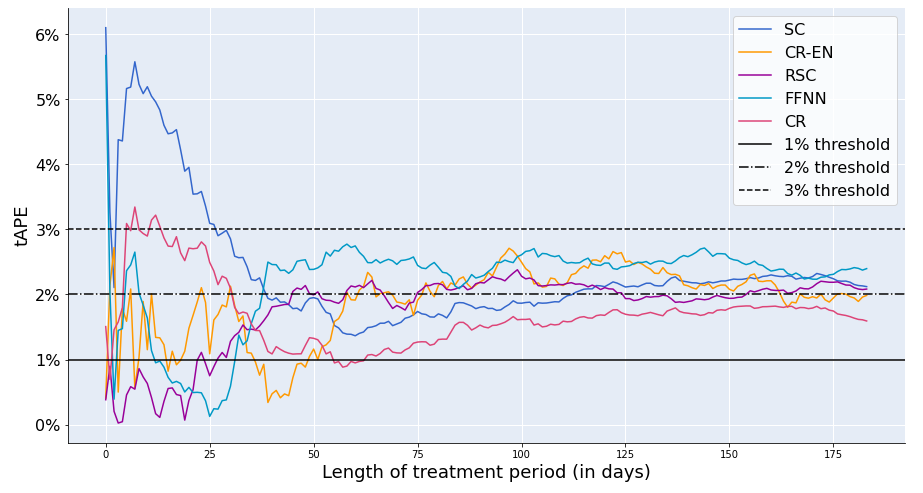}
		\caption{$S1$}
	\end{subfigure}
	\begin{subfigure}[b]{0.5\linewidth}
		\includegraphics[width=\linewidth]{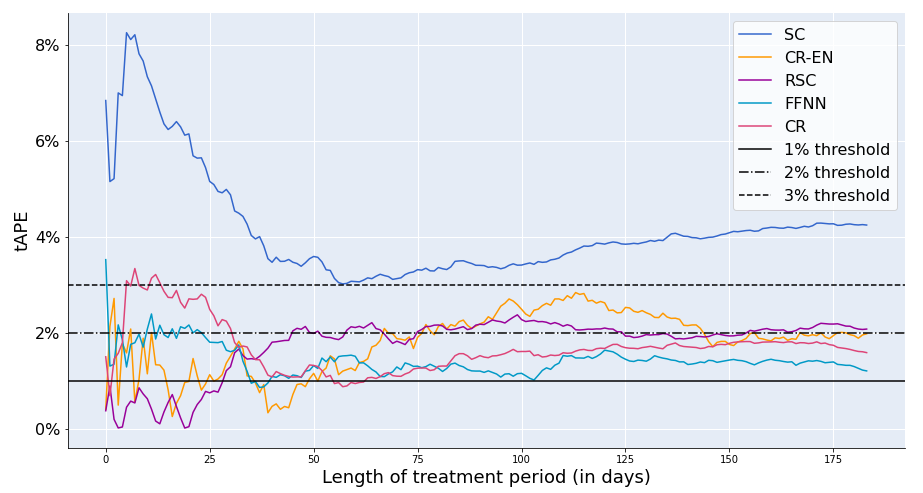}
		\caption{$S2$}
	\end{subfigure}
	\caption{Values of tAPE varying with the length of the treatment period for pseudo-treatment period 10}
\end{figure} 

\begin{figure}[H]
	\begin{subfigure}[b]{0.5\linewidth}
		\includegraphics[width=\linewidth]{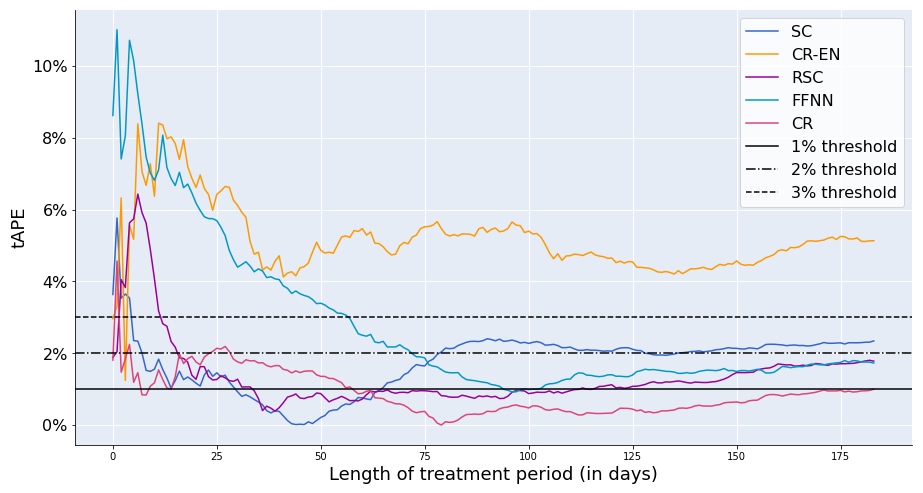}
		\caption{$S1$}
	\end{subfigure}
	\begin{subfigure}[b]{0.5\linewidth}
		\includegraphics[width=\linewidth]{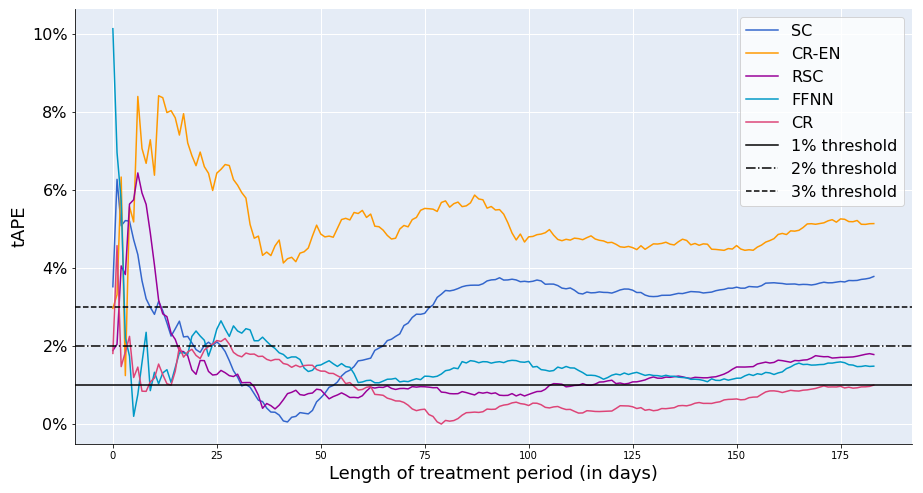}
		\caption{$S2$}
	\end{subfigure}
	\caption{Values of tAPE varying with the length of the treatment period for pseudo-treatment period 11}
\end{figure} 

\begin{figure}[H]
	\begin{subfigure}[b]{0.5\linewidth}
		\includegraphics[width=\linewidth]{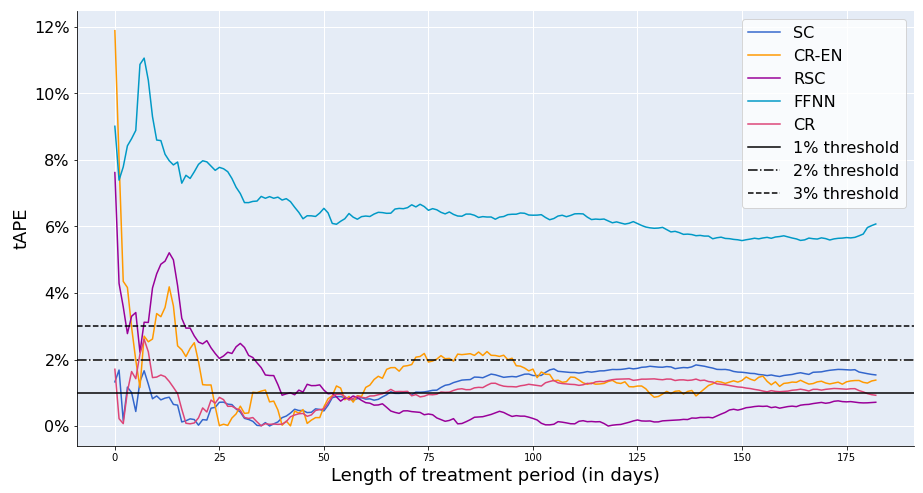}
		\caption{$S1$}
	\end{subfigure}
	\begin{subfigure}[b]{0.5\linewidth}
		\includegraphics[width=\linewidth]{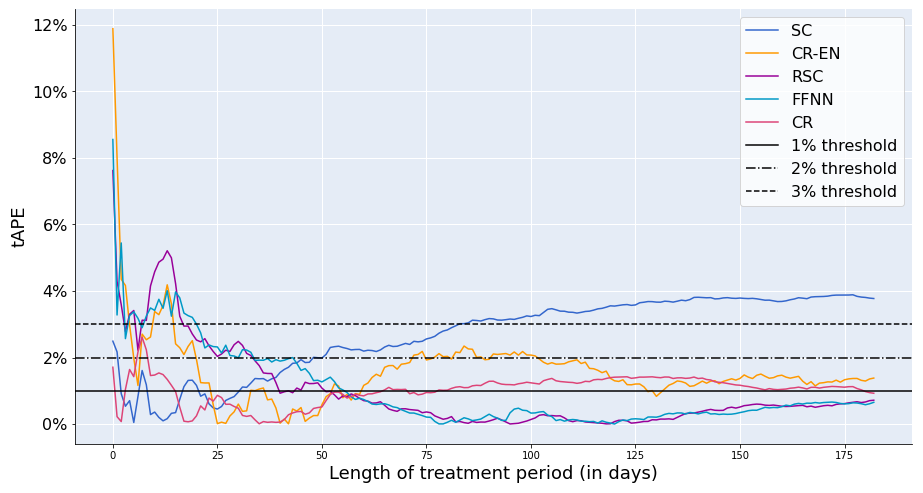}
		\caption{$S2$}
	\end{subfigure}
	\caption{Values of tAPE varying with the length of the treatment period for pseudo-treatment period 12}
\end{figure} 

\begin{figure}[H]
	\begin{subfigure}[b]{0.5\linewidth}
		\includegraphics[width=\linewidth]{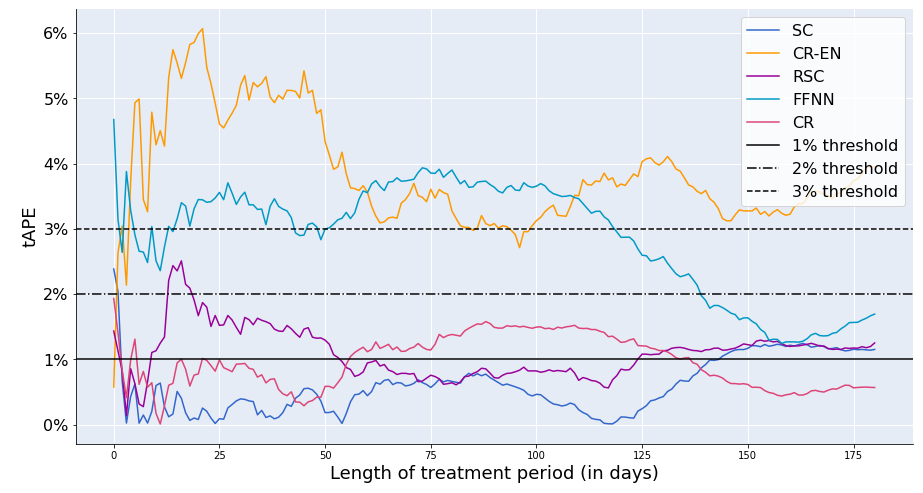}
		\caption{$S1$}
	\end{subfigure}
	\begin{subfigure}[b]{0.5\linewidth}
		\includegraphics[width=\linewidth]{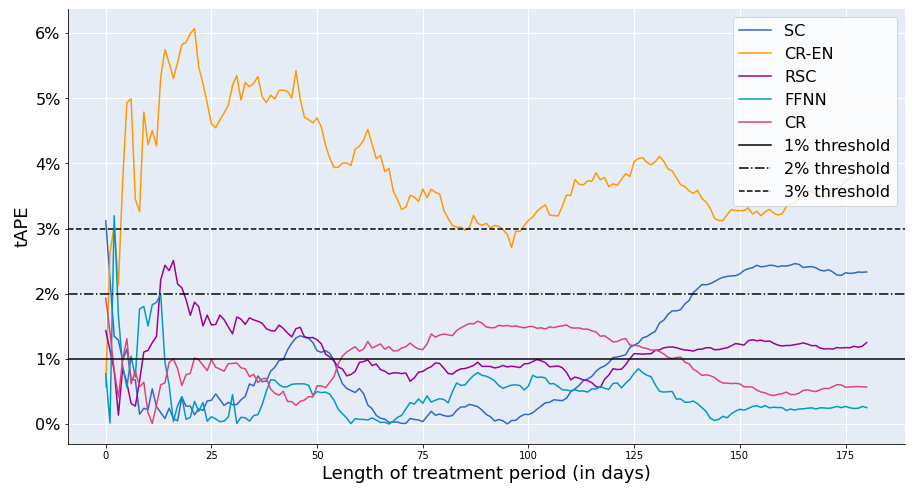}
		\caption{$S2$}
	\end{subfigure}
	\caption{Values of tAPE varying with the length of the treatment period for pseudo-treatment period 13}
\end{figure} 

\begin{figure}[H]
	\begin{subfigure}[b]{0.5\linewidth}
		\includegraphics[width=\linewidth]{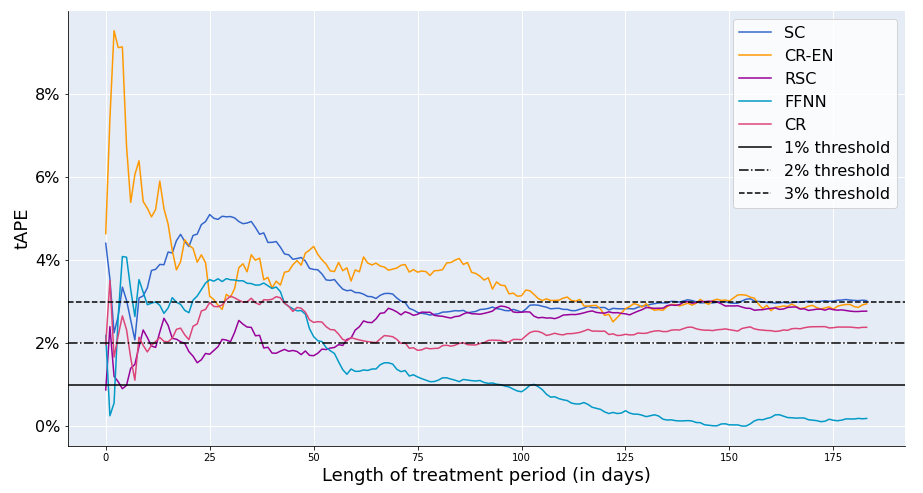}
		\caption{$S1$}
	\end{subfigure}
	\begin{subfigure}[b]{0.5\linewidth}
		\includegraphics[width=\linewidth]{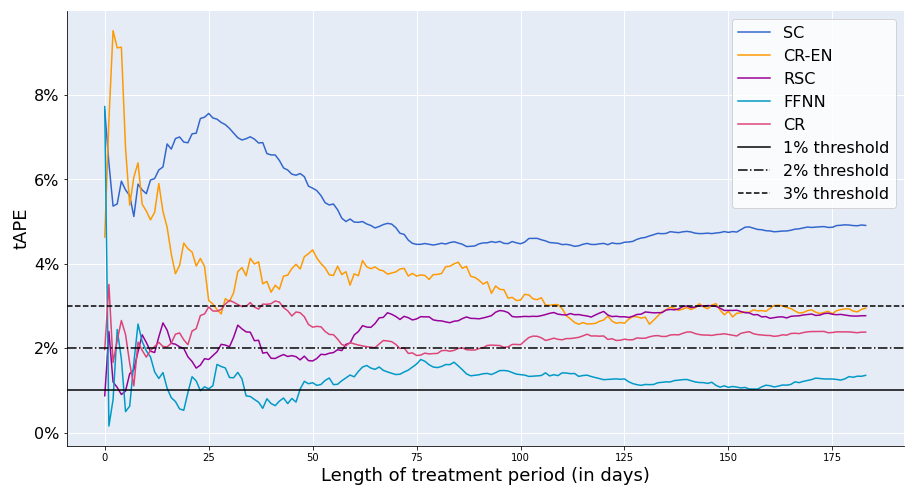}
		\caption{$S2$}
	\end{subfigure}
	\caption{Values of tAPE varying with the length of the treatment period for pseudo-treatment period 14}
\end{figure} 

\begin{figure}[H]
	\begin{subfigure}[b]{0.5\linewidth}
		\includegraphics[width=\linewidth]{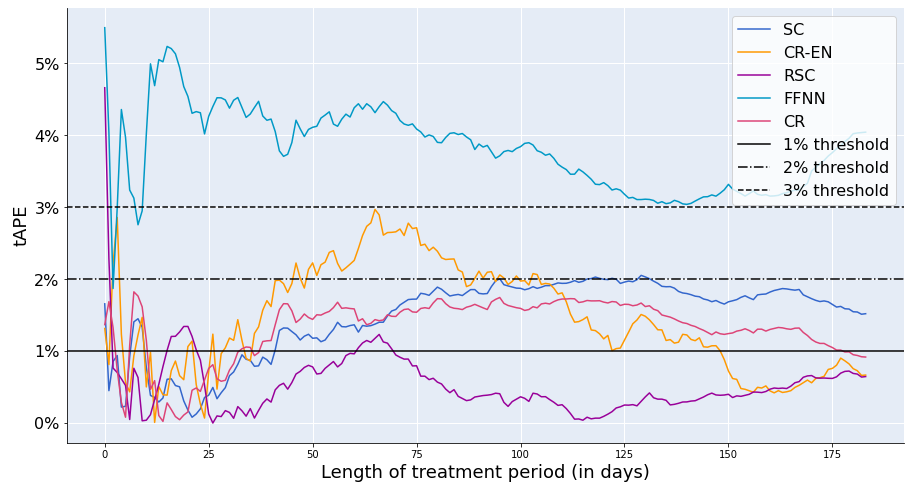}
		\caption{$S1$}
	\end{subfigure}
	\begin{subfigure}[b]{0.5\linewidth}
		\includegraphics[width=\linewidth]{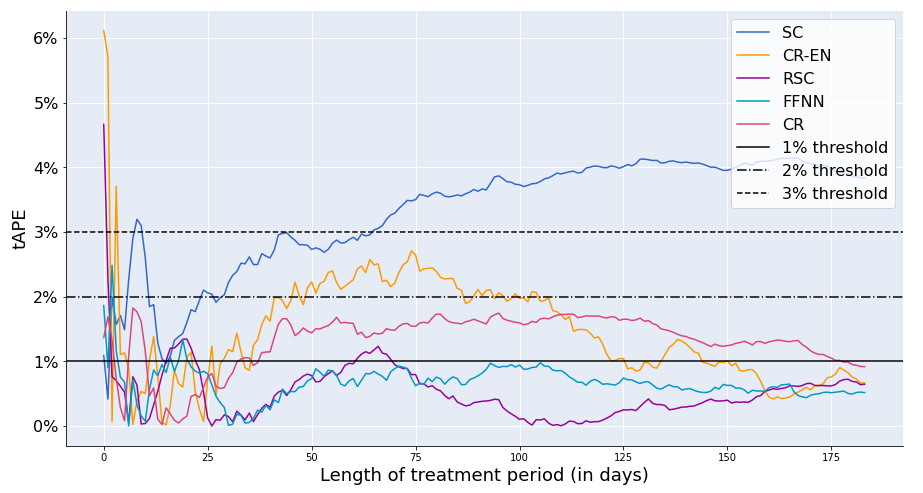}
		\caption{$S2$}
	\end{subfigure}
	\caption{Values of tAPE varying with the length of the treatment period for pseudo-treatment period 15}
\end{figure} 

\end{document}